%% file: main.tex
\pdfoutput=1
\documentclass[11pt]{article}

\usepackage[]{acl}

\usepackage{times}
\usepackage{latexsym}
\usepackage{amssymb}
\usepackage{pgfplots}
\usepackage{scalefnt}
\usepackage{boldline}
\usepackage{xcolor}
\usepackage{multirow}
\usepackage{makecell}
\usepackage{graphicx}
\usepackage{bm}
\usepackage{amsmath}
\usepackage{bbm}
\usepackage{float}
\usepackage{stfloats}
\usepackage{lipsum}
\usepackage{comment}
\usepackage{booktabs}
\usepackage{subfig}
\usepackage{lipsum}
\usepackage{cleveref}

\usepackage[flushleft]{threeparttable}

\crefformat{section}{\S#2#1#3} 
\crefformat{subsection}{\S#2#1#3}

\usepackage{booktabs}
\usepackage[T1]{fontenc}
\usepackage[utf8]{inputenc}

\usepackage{microtype}

\usepackage{inconsolata}

\usepackage{tabularx}

\usepackage{xspace}

\usepackage{hyperref}
\usepackage{xcolor}
\usepackage{listings}
\usepackage{color}
\usepackage{xspace}
\usepackage{multirow}
\definecolor{backcolour}{rgb}{0.95,0.95,0.92}
\lstdefinestyle{data_prompt}{
    basicstyle=\rmfamily\scriptsize,
    breakatwhitespace=true,
    breaklines=true,
    showstringspaces=true
}
\lstdefinestyle{exp_prompt}{
    basicstyle=\rmfamily\footnotesize,
    breaklines=true,
    breakatwhitespace=true
}
\input{commands}
\newcommand{\OURS}{Knowledge-Augmented Phrase Retrieval\xspace}
\newcommand{\task}{\textsc{Ktrl+F}\xspace}
\newcommand{\TASK}{knowledge-augmented in-document search\xspace}
\newcommand{\ctrlf}{Ctrl+F\xspace}
\newcommand{\reqo}{\textsc{req 1}\xspace}
\newcommand{\reqtw}{\textsc{req 2}\xspace}
\newcommand{\reqth}{\textsc{req 3}\xspace}
\title{\task: Knowledge-Augmented In-Document Search}

\author{
    Hanseok Oh$^1${\thanks{\, \* indicates equal contribution}}
        \quad
    Haebin Shin$^{1,2*}$
        \quad
    Miyoung Ko$^1$
        \quad
    Hyunji Lee$^1$
        \quad
    Minjoon Seo$^1$ \\
    $^1$KAIST AI \quad $^2$Samsung Research \\
    \texttt{\{hanseok, haebin.shin, miyoungko, hyunji.amy.lee, minjoon\}@kaist.ac.kr} 
}

\begin{document}
\maketitle

\begin{abstract}
    \input{KtrlF/1_abstract}
\end{abstract}

\section{Introduction}
\label{sec:intro}
\input{KtrlF/2_introduction}

\section{Related Works}
\label{sec:related_work}
\input{KtrlF/3_relatedWork}

\section{Ktrl+F: Knowledge-Augmented In-Document Search}
\label{sec:task}
\input{KtrlF/4_ktrlF}

\section{\task Dataset}
\label{sec:data}
\input{KtrlF/5_data}

\section{\OURS}
\label{sec:method}
\input{KtrlF/6_baselines}

\section{Experiments}
\label{sec:experiment}
\input{KtrlF/7_experiment}

\section{User Study}
\label{sec:user_study}
\input{KtrlF/8_user_study}

\section{Conclusion \& Future Work}
\label{sec:conclusion}
\input{KtrlF/9_conclusion}

\section*{Limitations}
% \section{Limitations}
\label{sec:limitations}
\input{KtrlF/a_limitations}

\section*{Acknowledgements}
\label{sec:acknowledgements}
\input{KtrlF/a_acknowledgments}

\bibliographystyle{acl_natbib}
\bibliography{anthology,custom}

\clearpage
\newpage
\appendix

% \section{Appendix}
\label{sec:appendix}
\input{KtrlF/a_appendix}

\end{document}

%% file: commands.tex
\definecolor{purple1}{HTML}{7570b3}
\definecolor{green1}{HTML}{1b9e77}
\definecolor{orange}{HTML}{d95f02}
\definecolor{green2}{HTML}{4daf4a}
\definecolor{red}{HTML}{e41a1c}
\definecolor{blue}{HTML}{377eb8}
\definecolor{purple2}{HTML}{984ea3}

%% file: KtrlF/1_abstract.tex
% Abstract
We introduce a new problem \task, a \TASK that necessitates real-time identification of all semantic targets within a document with the awareness of external sources through a single natural query. 
\task addresses following unique challenges for in-document search: 1) utilizing knowledge outside the document for extended use of additional information about targets, and 2) balancing between real-time applicability with the performance.
We analyze various baselines in \task and find limitations of existing models, such as hallucinations, high latency, or difficulties in leveraging external knowledge.
Therefore, we propose a \OURS model that shows a promising balance between speed and performance by simply augmenting external knowledge in phrase embedding. 
We also conduct a user study to verify whether solving \task can enhance search experience for users.
It demonstrates that even with our simple model, users can reduce the time for searching with less queries and reduced extra visits to other sources for collecting evidence.  
We encourage the research community to work on \task to enhance more efficient in-document information access.\footnote{Code, Chrome extension plugin, and dataset are available at \url{https://github.com/kaistAI/KtrlF}}

%% file: KtrlF/2_introduction.tex
% Introduction
% % overlooked problem & still people use CtrlF
Despite significant advancement in many Natural Language Processing applications, facilitated by transformer-based models ~\citep{devlin2018bert,raffel2019t5}, real-time in-document search still leans heavily on conventional lexical matching tools like the "Find" function (\ctrlf) and regular expressions.
These tools, while fast, have clear limitations, especially with ambiguous keywords or multiple targets.

\begin{figure*}[t!]
\small
\centering
\includegraphics[width=1.0\textwidth]{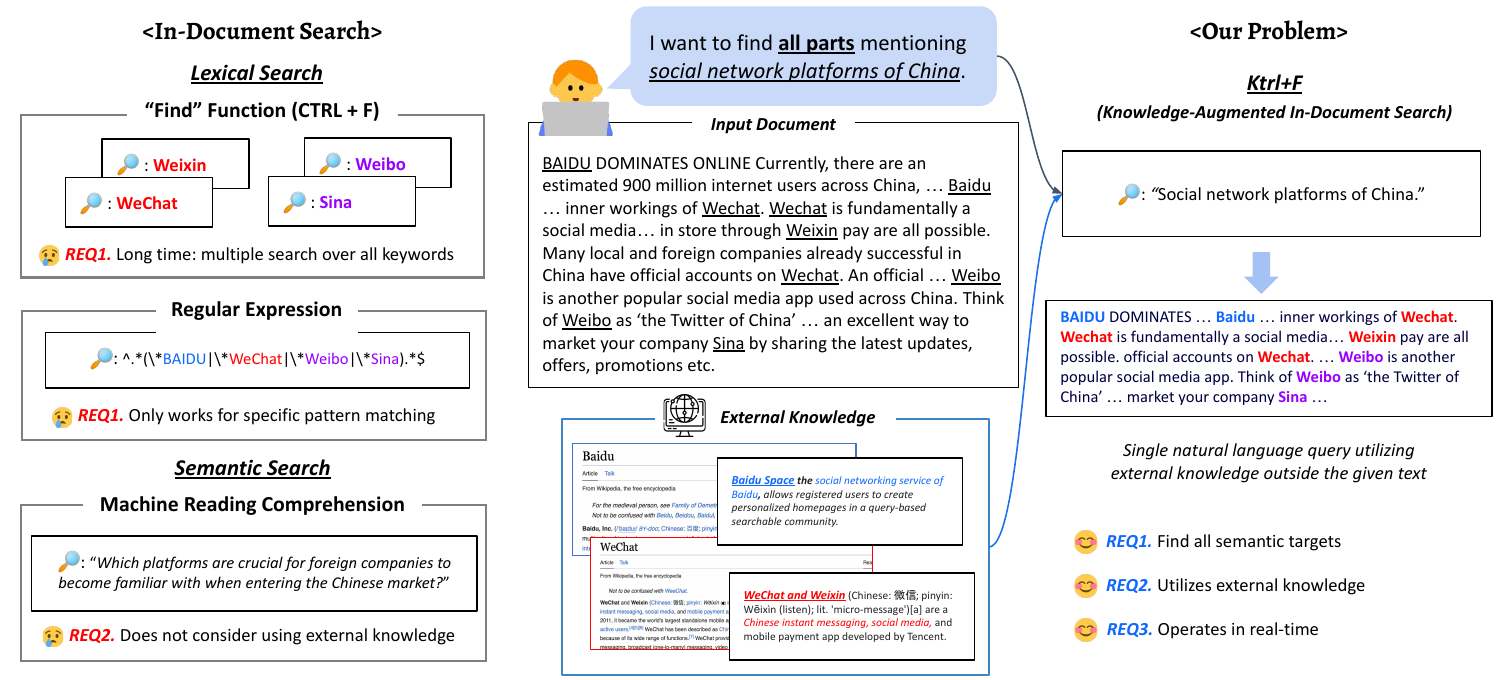}
\caption{Comparison between in-document search and \task problem. 
In-document search accesses the information in documents by either lexical search (\ctrlf, Regular expression) or semantic search (MRC). 
Lexical search suffers from finding semantically matching keywords, and semantic search does not consider external knowledge. 
\task requires an efficient way to utilize external knowledge to find all semantic targets in real-time. }
\label{fig: system}
% \vspace{-0.5em}
\end{figure*}

Machine Reading Comprehension (MRC) seems a promising solution to these issues. It reads documents, comprehends their context, and answers questions~\citep{rajpurkar2016squad}.
However, MRC focuses on explicit contents, limiting its value when users need knowledge not directly in the document~\citep{trischler-etal-2017-newsqa,rajpurkar-etal-2018-know, joshi-etal-2017-triviaqa}.
Consider a scenario where users read a news article and seek for information on the \textit{"Social network platform of China."} (Figure~\ref{fig: system}).
Typically, users refer to external sources such as Wikipedia to gather additional details not explicitly mentioned in news related to candidates, such as \textit{WeChat, Baidu, and Twitter}.
An alternative is harnessing the capabilities of powerful pre-trained language models~\citep{brown2020language,touvron2023llama2}.
However, their generative nature poses challenges for real-time search task.

% para2) introduce Ktrl+F
To overcome the limitations of previous methods and enhance the efficiency and comprehensiveness of in-document search, we present a new problem ~\task~(\TASK).
This task aims to reduce redundancy and better meet the requirements of real users.
Given a natural language query and a long input document, \task is designed to fulfill three key criteria:
(\reqo) Find all semantic targets. 
(\reqtw) Utilizes external knowledge. 
(\reqth) Operates in real-time.
In the absence of a suitable dataset to evaluate \task, we curate a new dataset with unique queries demanding matching external evidence.
To measure model performance in \task, we introduce a set of reformulated metrics tailored to measure processing speed while maintaining robust and high performance.

% para3) baseline & suggest simple model for KTRL+F
We conduct an extensive analysis of various baselines for \task and find several limitations including hallucination, slow speed with generative models, and challenges in incorporating external knowledge into MRC models (see \cref{sec:experiment-analysis} for details).
To strike a balance between real-time processing speed and achieving high performance through effective utilization of additional knowledge, we introduce a simple yet effective extension of phrase retrieval~\citep{lee2021learning}: \OURS.
This model seamlessly extends the phrase retrieval to cater to in-document search scenarios, all while integrating external knowledge without the need for additional training steps.
Our experiments support that by simply adding the knowledge embedding and the phrase embedding, \OURS exhibits the potential to reflect external knowledge without sacrificing latency.

% User study
Furthermore, we conduct a user study to show the necessity of \task utilizing a Chrome extension plugin that operates in the real web environments, built upon our model.
Results of the study demonstrate that search experience of users can be enhanced even with our simple model with seamless access to external knowledge during in-document searches.
We encourage the research community to take on the unique challenge of solving \task requiring balance between performance and speed to enhance more efficient and effective information access.

%% file: KtrlF/3_relatedWork.tex
\paragraph{Machine Reading Comprehension (MRC)}
is a task to find the answer to a question in the provided context.
Most MRC datasets assess the ability of context understanding of the model by extracting a single span for the query only grounding on the information within a provided context \citep{rajpurkar2016squad,trischler-etal-2017-newsqa, joshi-etal-2017-triviaqa, rajpurkar-etal-2018-know,fisch-etal-2019-mrqa,kwiatkowski-etal-2019-natural}.
Few works explore the identification of multiple targets for a query in the input document evaluating the model's comprehension of the given context ~\citep{Dasigi2019QuorefAR,Zhu2020QuestionAW,li2022multispanqa} .
Some studies tackle information-seeking problem by utilizing external information missing from input document to gap knowledge ~\citep{ferguson2020iirc,Dasigi2021ADO}. 
This external information aids in enhancing the understanding of the context.
However, since the \task relies on external knowledge beyond its context, it is essential to explicitly ground external knowledge about the target. 
Consequently, the evaluation of \task focuses not on the understanding of the given context, but on information obtained from outside the given context.

\paragraph{Knowledge-augmented information retrieval}
is an approach to enrich external information within the text embedding.
% Why it is studied?
The introduction of a knowledge-augmented design aims to supplement deficient contextual information, thereby enhancing the richness of text embedding.
Numerous studies tackle knowledge augmentation across various NLP tasks \citep{zhang2019ernie,xiong2019pretrained, peters2019knowledge, poerner2020bert, fevry2020entities, levine2020sensebert,wang2021k,bertsch2023unlimiformer}.
% Message & Why
The integration of information from diverse sources leads to an improved language understanding ability.
% Knowledge Augmentation in IR task 
However, the application of knowledge augmentation in information retrieval tasks has received comparatively less attention.
\citet{Lin2022AggretrieverAS} attempts to improve text embeddings for retrieval by enriching context information through embeddings derived from a given context, without specifically focusing on external knowledge.
Meanwhile, \citet{Lee2022ContextualizedGR} utilizes contextualized embeddings as vocabulary embeddings for text tokens in a generative retriever, thereby enhancing contextual information for basic text tokens. 
Additionally, \citet{Raina2023ERATEER} focuses on the retrieval augmented text embedding to efficiently reuse prebuilt dense representation with lightweight representation, and also discusses the necessity of systems for utilizing external contextual information to include contextual information outside the given context in text embedding tasks. 
% Connection with our work 
In contrast to these approaches, \task directs its attention on augmenting knowledge from external sources for entities in a novel in-document retrieval task.
This involves extracting information not present in the given text, thus expanding the capabilities of the information retrieval process.

%% file: KtrlF/4_ktrlF.tex
In this section, we define \task, which is \TASK task and its unique characteristics~(\cref{sec:3.1_problem_definition}).
Then we describe the evaluation metrics to measure each requirement~(\cref{sec:3.2_evluation_metrics}).
\subsection{Task Definition}
\label{sec:3.1_problem_definition}
\task is a task that requires finding all semantic targets from a given input document in real-time with the awareness of external knowledge, when given a natural language query.
As illustrated in Figure ~\ref{fig: system}, when presented with a natural language query and a input document, Ktrl+F is designed to meet three essential criteria.

\paragraph{\reqo: Find all semantic targets.}
\task requires finding all relevant targets within a given document. The term "all" refers to multiple aspects: finding all multiple answers (\textcolor{blue}{baidu}, \textcolor{red}{wechat}, \textcolor{violet}{weibo}), all occurrences of each answer (\textcolor{blue}{baidu} appears two times in the document), and all lexical variations of mentions for each answer (\textcolor{violet}{Weibo}, \textcolor{violet}{Sina}).

\paragraph{\reqtw: Utilize external knowledge.}
Expanding the matching space from lexical to semantic introduces a comprehensive connection between query and target units.
However, in many cases, targets contain extra information beyond the input document. 
By effectively leveraging this additional information through utilization of external knowledge, we can further bridge the semantic gap between the query and the targets.

\paragraph{\reqth: Search in real-time.}
\task inherits the practicality of in-document search, such as \ctrlf, which emphasizes real-time search to minimize the time on finding targets within the input document.
The complexity of \task lies in effectively balancing real-time applicability with the performance of finding all matching targets by leveraging external knowledge.

\subsection{Evaluation Metrics}
\label{sec:3.2_evluation_metrics}

To assess various aspects of \task, we employ a range of metrics that collectively measure the overall balance of performance and speed.
Following \citet{izacard-grave-2021-leveraging}, we indirectly assess the impact of utilizing external knowledge by comparing the overall performance of the system with and without its incorporation, given the absence of a definite gold standard answer (\reqtw).

\paragraph{List EM F1, List Overlap F1, Robustness Score.}
The three metrics measure if the model finds all semantic targets, which fulfills \reqo.
% List EM considers correct only when the prediction list is exactly the same as the ground truth list. Whereas, List Overlap allows partial matches between individual elements of the predicted and the ground truth list, as described in Equation~\ref{equation:list_overlap}. 
List EM considers correct only when the prediction list is exactly the same as the ground truth list. Note that List EM is different from Set EM, a commonly used metric in Machine Reading Comprehension~\cite{rajpurkar2016squad}, in that List EM aims to identify all occurrences of targets within a input document. Whereas, List Overlap allows partial matches between individual elements of the predicted and the ground truth list, extending set-based partial match from MultispanQA~\cite{li2022multispanqa}. For detailed equations and explanation for List Overlap, please refer to Appendix~\ref{appendix:list_overlap_metric}.

Inspired by \citet{zhong2023romqa}, we adjust robustness score to assess the robustness of system in predicting target answer entities as queries change within a given input document. 
Treating queries linked to the same document as a cohesive cluster, we calculate the robustness score by averaging the minimum score within each cluster.
This approach enhances the comprehensive evaluation of \task task, given that the knowledge-augmented design of \task allows for various queries with different target answers for in-document searches.

\paragraph{Latency.}
Latency is a metric for assessing real-time applicability, therefore satisfying \reqth.
We measure in ms/Q (millisecond per query) which is widely used in retrieval systems to represent query inference speed~\citep{khattab2020colbert, santhanam-etal-2022-colbertv2}.

%% file: KtrlF/5_data.tex
% \begin{figure}[t!]
\begin{figure*}[t!]
% \begin{minipage}{\columnwidth}
\small
\centering
\includegraphics[width=1.0\textwidth]{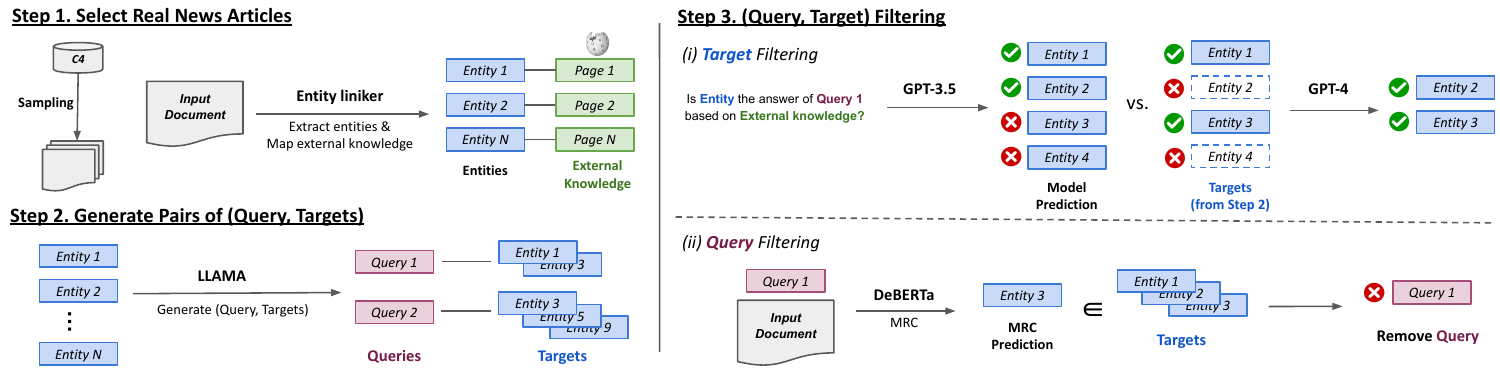}
\caption{Overview of \task dataset construction pipeline. We utilize real news articles as input documents (Step 1), and automatically generate queries and targets using LLAMA (Step 2). 
% To improve the reliability of targets, we identify whether each entity is a target by referring to external knowledge, and finalize entities that differ from the Step 2 results (Step 3-1).
To enhance the reliability of the identified targets, each entity is re-verified with external knowledge and finalized in (Step 3-1). 
% Additionally, we remove queries that do not satisfy \reqtw using the MRC model (Step 3-2).
Additionally, we use the MRC model to eliminate queries that do not meet the criteria outlined in \reqtw (Step 3-2). 
}
\label{fig: data_pipeline}
% \end{minipage}
% \end{figure}
\end{figure*}

\label{sec:3.3_dataset}
We introduce a data construction pipeline to assemble essential components of \task: input document, query, corresponding targets, and external knowledge (Figure~\ref{fig: data_pipeline}).
Then we describe human verification procedures to ensure quality. 

\subsection{Dataset Construction Pipeline}
\paragraph{Step 1. Select Real News Articles.}
To simulate real-world document scenarios, we randomly sample 100 English news articles from the publicly available C4~\citep{raffel2019t5} after preprocessing them based on their length and the number of entities. 
We utilize an entity linking API\footnote{\url{https://cloud.google.com/natural-language/docs/analyzing-entities}} to identify all entities within the article and extract external knowledge (i.e., Wikipedia) linked to the entities. 
Details of preprocessing and external knowledge are described in the Appendix \ref{appendix:dataset_details}.

\paragraph{Step 2. Generate Pairs of (Query, Targets).}
Using the entities extracted from each input document (Step 1), we utilize LLAMA-2-Chat-70B~\citep{touvron2023llama2} to generate diverse queries and targets (prompt in Figure~\ref{fig:query_generation_prompt}). 
We generate 10 questions for each input document. 
To satisfy the criteria of utilizing external knowledge (\reqtw), we provide only the extracted entities into the model, excluding the input document.
This is done to remove the dependency on the document itself, as \task prioritizes queries that cannot be answered solely with the document and requires the integration of external knowledge.

\paragraph{Step 3-1. Target Filtering.}
To mitigate the potential problem of false positive and false negative in the generated targets by LLAMA-2-Chat-70B~\citep{touvron2023llama2}, we implement an additional process inspired by ~\citet{zhong2023romqa}. This process determines whether each entity is the answer to the query, leveraging external knowledge (prompt in Figure~\ref{fig:answer_refine_prompt}).
Initially, we utilize GPT-3.5 (\texttt{gpt-3.5-turbo-0613})~\citep{openai2022chatgpt} to identify entities judged as potential answer targets. Subsequently, GPT-4 (\texttt{gpt-4-0613})~\citep{openai2023gpt4} makes the final decision for entities where there is a disagreement between GPT-3.5 and the results of Step 2.
Detailed statistics of the results by each model are available in the Appendix~\ref{appendix:dataset_details}.

\paragraph{Step 3-2. Query Filtering.}
Though we prioritize queries that require integrating external knowledge in Step 2, there are still many queries that do not meet \reqtw. To further reduce the number of such queries, we utilize a DeBERTaV3-large~\citep{he2023debertav3}\footnote{\url{https://huggingface.co/deepset/deberta-v3-large-squad2}}, finetuned using the SQuAD 2.0~\citep{rajpurkar-etal-2018-know}. We specifically exclude queries that the MRC model can answer solely based on the input document, leaving only suitable queries for \reqtw.
Finally, 512 queries are collected out of the 1,000 queries generated in Step 2.
See Appendix~\ref{appendix:dataset_details} for detailed scoring criteria of the MRC model.

\subsection{Dataset Analysis}

\paragraph{Human verification setup.}
To assess the quality of the auto-generated dataset, we conduct human verification on a randomly selected subset of 104 queries, representing about 20\% of the entire dataset. 
Eight annotators participated, with three assigned to evaluate each sample to minimize personal bias.
Annotators are tasked with responding to three specific questions: two for query-side verification (\textit{Q1} and \textit{Q2}) and one for target-side verification (\textit{Q3}).

The first question (\textit{Q1}) assesses how well the generated query aligns with \reqtw. 
Annotators identify evidence for each target to answer the query, with the ideal response being annotators stating that evidence cannot be found in the input document for all targets.
The second question (\textit{Q2}) evaluates the naturalness of the generated query by choosing the type of unnatural query: \textit{"Ambivalent or subjective expressions"}, \textit{"Lack of factual basis"}, \textit{"Logical errors"}, \textit{"etc"}. The ideal response is for annotators to select \textit{"None of these options"}, indicating a naturalness in the generated queries.
The third question (\textit{Q3}) focuses on evaluating the reliability of auto-generated targets. 
Annotators select the correct target for the query by referring to Wikipedia, mirroring the process in target filtering (Step 3-1) in the dataset construction pipeline. This establishes the reliability between the annotator's response and the dataset. 
Target-side verification is conducted on a distinct set of 104 samples from query-side verification.
The user interface and detailed instructions for each question are presented in Figure~\ref{fig:human_verification_page}.

\begin{table}[t]
\begin{minipage}{\columnwidth}
\centering
\resizebox{1.0\textwidth}{!}{
    \begin{tabular}{lr}
    \toprule
     \textit{Q1. Is it possible to answer using only the input doc?} & \\
    \midrule
     \hspace{1.0em}Need more external knowledge & 74.3\% \\
     \hspace{1.0em}Don't need external knowledge & 25.7\% \\
     \hspace{2.0em}\% of answered targets & 43.6\% \\
    \toprule
     \textit{Q2. Is it unnatural query?} & \\
    \midrule
     \hspace{1.0em}Natural Query & 95.0\% \\
     \hspace{1.0em}Subjective Query & 3.0\% \\
     \hspace{1.0em}etc. & 2.0\% \\
    \toprule
     \textit{Q3. Reliability of Target determination} & \\
    \midrule
     \hspace{1.0em}kappa coefficient ($\kappa$) & 0.627 \\
    \bottomrule
    \end{tabular}
}
\caption{Human Verification Results}
\label{tab:human_verification_results}
\end{minipage}
\end{table}

\begin{table}[t]
\begin{minipage}{\columnwidth}
\centering
\resizebox{1.0\textwidth}{!}{
    \begin{tabular}{lcccc}
    \toprule
      & Avg. & Min. & Max. \\
    \midrule
     Length of Input Document & 1974 & 999 & 3254 \\
     % \midrule
     Queries per Input Document & 5.2 & 1 & 10 \\
     % \midrule
     Answer Mentions per Query & 4.2 & 1 & 30 \\
     Answer Entities per Query & 1.8 & 1 & 8 \\
     \bottomrule
    \end{tabular}
}
\caption{Statistics of \task Dataset}
\label{tab:evaluation_set_statistics}
\end{minipage}
% \vspace{-1em}
\end{table}

\paragraph{Dataset quality and statistics.}
Since all samples are evaluated by three annotators, final human judgment is determined through majority voting. 
The inter-annotator reliability is detailed in Appendix \ref{appendix:inter-rater_reliability}.
For the first question, 74.3\% of samples are considered unable to answer the target solely based on the input document. 
Of the remaining 25.7\% of samples, only 43.6\% of targets can be solved solely based on the input document.
This indicates that our auto-generated dataset is suitable for evaluating \task requiring additional knowledge beyond the semantic information present in the input document. 
About the naturalness of query (Q2), 95\% of samples are considered natural, while 3\% are subjective. 
About 2\% of the samples contain unnatural queries for other reasons, such as entities being directly mentioned in the query.
For the third question, we find a kappa coefficient~\citep{Cohen1960ACO} of $\kappa=0.627$ between humans and the dataset.
Following \citet{Landis1977TheMO}, this indicates \textit{substantial agreement} between human judgment and the data construction pipeline.
In total, the \task dataset comprises 512 queries for 98 input documents with an average of 4.2 mentions per query (Table~\ref{tab:evaluation_set_statistics}). More examples of the \task dataset are available in Table~\ref{table:appendix_data_example}.

%% file: KtrlF/6_baselines.tex
\label{sec}

\begin{figure}[t!]
\begin{minipage}{\columnwidth}
\small
\centering
\includegraphics[width=1.0\textwidth]{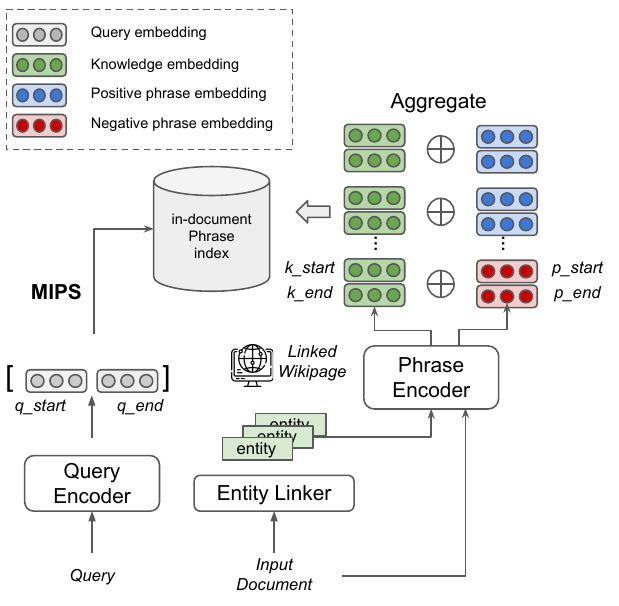}
\caption{Overview of \OURS. 
}
\label{fig: method}
\end{minipage}
% \vspace{-1em}
\end{figure}

The challenge of \task is to effectively balance real-time applicability and high performance while utilizing efficient use of external knowledge.
To meet the three requirements of \task, we propose \OURS extending the phrase retrieval architecture of DensePhrases~\citep{lee2021learning} within the setting of in-document search and enriching external knowledge about potential targets with external knowledge linking and knowledge aggregation modules as illustrated in Figure~\ref{fig: method}.
Notably, our model doesn't require an additional training step.

\subsection{External Knowledge Linking Module}
The external knowledge linking module scans the target text, identifies entities that could be potential targets, and maps each of them to the relevant Wikipedia knowledge base. 
The module outputs a list of candidate targets along with the linked Wikipedia page for each target, serving as external knowledge about the targets.
We use existing entity-liners to focus on building models that can integrate external knowledge.
While there are various entity-linkers available, we choose to utilize a Wikifier API~\citep{brank2017annotating} as an entity linker for its ease of use.

\subsection{Query and Phrase Encoder}
The phrase and query encoder modules handle the encoding of the candidate phrase and the query, respectively. 
We utilize the pre-trained DensePhrases model~\citep{lee2021learning} to extract phrase embeddings.
For the query embedding, we extract the special token \texttt{[CLS]} from the output embeddings of the query encoder.
We use two distinct query encoders to extract the start and end position embeddings for the query, following \citet{lee2021learning}.
Subsequently, we concatenate the corresponding token embeddings, denoted as $[q_{start}; q_{end}] \in \mathbb{R}^{2d}$, to create a query embedding.
Similarly, for the phrase encoder, we use concatenated token-level embeddings of the entity's boundary tokens (start and end token embeddings denoted as $[p_{start}, p_{end}]$) as the phrase embedding.

\subsection{Knowledge Aggregation Module}
To integrate external knowledge related to the entity, we employ the same phrase encoder used for extracting embeddings for candidate entities. 
Following the approach in \citet{Lee2022ContextualizedGR}, we generate a knowledge embedding, denoted as $[k_{start}; k_{end}] \in \mathbb{R}^{2d}$, for the linked entity by concatenating the name of entity and its corresponding Wikipedia page (refer to Figure~\ref{fig: method_external_embedding} for details).
This effectively encodes relevant knowledge about the entity into its embedding.
To combine external knowledge embedding with the entity embedding and create an in-document phrase index, we use a straightforward element-wise addition operation. 
This demonstrates promising results in our experiments enabling the system to capture the contextual knowledge for more accurate and comprehensive search and retrieval within the document without requiring further tuning.
Through the Maximum Inner Product Search (MIPS) operation, \OURS can identify all matching targets in real time.

%% file: KtrlF/7_experiment.tex
% Experiments
\begin{table*}[t!]
\begin{minipage}{\textwidth}
\centering
\resizebox{1.0\textwidth}{!}{
    \begin{tabular}{ccccccc}
    \toprule
     & & \multicolumn{1}{c}{Speed} & \multicolumn{4}{c}{Performance} \\     
     \cmidrule(lr){3-3} 
     \cmidrule(lr){4-7}
    Type & Model & Latency (ms/Q) ($\downarrow$) & List EM ($\uparrow$) & (R) List EM ($\uparrow$) & List Overlap ($\uparrow$)& (R) List Overlap ($\uparrow$) \\
     \midrule
     \multirow{6}{*}{Generative}
     & GPT-3.5 & - & \underline{30.346} & \underline{8.284} & \textbf{41.929} & 19.446\\
     & GPT-4 &  - & \textbf{30.457} & 7.452 & 37.402 & 12.898\\
     & LLAMA-2-Chat-7B & 2359 & 28.529 & \textbf{8.947} & 40.546 & 20.008\\
     & LLAMA-2-Chat-13B & 3176 & 28.846 & 8.024 & 37.098 & 14.367\\
     & VICUNA-7B-v1.5 &  1951 & 17.831 & 3.694 & 31.216 & 12.532\\
     & VICUNA-13B-v1.5 &  2420 & 24.490 & 6.977 & 39.278 & \underline{20.401}\\
     \midrule
     Extractive & SequenceTagger  & \underline{26} & 7.239 & 0.612 & 8.614 & 1.211 \\
     \midrule
     \multirow{2}{*}{Retrieval}
     & Ours (w/ Wikifier)  & \textbf{15} & 23.152 & 7.091 & \underline{40.718}  & \textbf{23.107}\\
     & Ours (w/ Gold) & 14 & 46.170 & 22.426 & 53.689 &  32.285 \\
     
     \bottomrule
    \end{tabular}
}
\caption{Speed and performance evaluation results for \task dataset.
Note that API-based models (GPT-3.5 and GPT-4) are excluded from speed evaluation. 
Robustness scores are noted with (R) with corresponding metric.
Ours denotes \OURS, and the best results excluding Ours (w/ Gold) are in bold, while second-best ones are underlined.}
\label{tab:main_result}
\end{minipage}
\end{table*}

\subsection{Setup}
\label{sec:experiment-setup}
% Update - final version : mention RAG baselines
When selecting baselines, our primary focus lies in evaluating the effectiveness of various representative options in addressing \task.
We categorize potential baseline types into generative, extractive, and retrieval (ours) models.

\paragraph{Generative baselines}
solve \task as a text generation problem, where the model takes instructions, a input text, and a query as input and sequentially produces matching targets (see Appendix \ref{appendix:generative_methods}).
The parametric space of Large Language Models (LLM) serves as an implicit source of general knowledge under the assumption that LLMs can serve as a closed-book model, as discussed by \citet{raffel2019t5,roberts-etal-2020-much,brown2020language,de2020autoregressive,yu2023generate}.
To explore the knowledge within the parametric space, we utilize various LLM models, such as the LLM API versions GPT-3.5~\citep{openai2022chatgpt} and GPT-4~\citep{openai2023gpt4}, as well as open-source models like LLAMA-2~\citep{touvron2023llama2} and VICUNA v1.5~\citep{vicuna2023}, ranging in size from 7B to 13B.
We additionally post-process generated outputs of models to only extract targets for evaluation.

Moreover, we observe that Retrieval Augmented Generation (RAG) baselines, which merely retrieve and enhance information from the query side, performs worse than naive LLM approaches. 
The unique characteristics of \task require grounding information from both the query and target text sides, presenting a distinct challenge.
Consequently, existing methods in the RAG models, which focus solely on retrieving knowledge from the query side,fail to adequately address this challenge. 
For detailed results and analysis of RAG baselines, please refer to Appendix \ref{appendix:rag_baseline}.

\paragraph{Extractive baseline}
is similar to extraction-based model for Machine Reading Comprehension task. 
This approach uses the internal knowledge within the target text to directly locate the answer spans. 
In order to find all relevant spans in the target text, we follow the previous works~\citep{segal2020simple, li2022multispanqa} that helps identify multiple entities. 
We utilize a BERT based sequence tagging model which is fine-tuned using MultiSpanQA~\citep{li2022multispanqa} dataset, denoted as SequenceTagger.

\begin{table*}[t!]
% \begin{minipage}{\columnwidth}
\centering
\resizebox{0.8\textwidth}{!}{
    \begin{tabular}{c|l|cccc}
    \toprule
     Entity Linker & \multicolumn{1}{c|}{Model} & List EM ($\uparrow$)&(R)List EM ($\uparrow$) & List Overlap ($\uparrow$) & (R)List Overlap ($\uparrow$)\\
     \midrule
     \multirow{3}{*}{\begin{tabular}{c} Gold \\(GCP API)\end{tabular}}
      & \textbf{Ours} & 46.170 & 22.426 & 53.689 & \textbf{32.285}\\
      & ~- External & 34.582 & 14.178 & 43.758 & 26.406 \\
      & ~- Internal & \textbf{47.345} & \textbf{23.097} & \textbf{54.308} & 30.599\\
     \midrule
     \multirow{3}{*}{Wikifier}
      & \textbf{Ours} & \textbf{23.152} & 7.091 & \textbf{40.718} & \textbf{23.107}\\
      & ~- External  & 15.620 & 4.742 & 31.805 & 18.823\\
      & ~- Internal & 22.851 & \textbf{7.773} & 39.391  & 20.812\\
     \bottomrule
    \end{tabular}
}
\caption{Ablation study on the impact of existence and quality of external knowledge. We measure the performance when using different entity linkers (Gold w/ GCP API, Wikifier API). 
We further evaluate the impact of contextual phrase embedding (Internal) and external embedding (External) by removing the related part.
}
\label{tab:entity_linker}
% \vspace{-0.5em}
\end{table*}

\subsection{Results}
\label{sec:experiment-analysis}
Lower latency means faster time to find targets\footnote{Speed measurements use an A6000 GPU on a server with two AMD EPYC 7513 CPUs, each with 32 physical cores.
}, and among various metrics, the List Overlap score can be indicative of general performance\footnote{We report the micro-averaged F1 scores. Detailed Precision and Recall scores are available in Table~\ref{tab:precision_recall}.}. Note that all models in the experiment are evaluated in a zero-shot manner. 
% \OURS with Wikifier records the best latency and GPT-3.5 shows best list overlap score among all. 

\paragraph{Generative and extractive baselines}
show difficulties in balancing real-time applicability and performance as 
Table~\ref{tab:main_result}.
GPT-3.5 excels in List Overlap scores, leveraging its parametric knowledge effectively. 
Interestingly, expanding model capacity doesn't consistently enhance performance unlike increasing latency.
Upon close examination of LLAMA-2 models, we can find possible reasons: smaller models generate more targets (avg. 3.347 for 7B, avg. 2.324 for 13B), leading to lower precision but higher recall, ultimately contributing to improved performance in List Overlap.
The generative nature of these models introduces complexities including challenges such as hallucination and difficulties in effective restriction of generated output (see examples in Table~\ref{table:hallucination}).
Conversely, the SequenceTagger, an extractive baseline, falls short in \task. 
Its inability to utilize external knowledge highlights the importance of incorporating such knowledge beyond the input document for successful \task resolution.
For a comprehensive baseline understanding, prediction example for each model is available in  Appendix~\ref{appendix:prediction_example} and additional experiments are reported in Appendix~\ref{appendix:additional_exeriments}. 
% \vspace{-2em}

\paragraph{\OURS}
demonstrates a balance between latency and achieving overall performance.
Incorporating knowledge embedding into the phrase retrieval process, our model (Ours w/ Wikifier) demonstrates competitive performance in List Overlap metrics, despite having a significantly smaller model capacity (330M, only 5\% of the smallest generative baseline) than other generative baselines.
When provided with gold entity linking information used in the dataset construction pipeline, our model achieves the best performance (Ours w/ Gold).
To compare with other baselines, we threshold the prediction results from top 4 according to the data distribution~\footnote{To provide a comprehensive understanding of the model, we additionally report MAP metrics in Table~\ref{tab:appendix-map} of Appendix~\ref{appendix:metrics}.}.
Beyond performance, the retrieval-based design of our model is suitable for real-time applicability, exhibiting smaller latency than other baselines. 
While our model demands extra time for the initial indexing of long input documents into searchable format, taking 2.863 and 0.955 seconds for our models with Wikifier and Gold respectively, the subsequent querying of the indexed text introduces real-time latency.
This shows a significant advantage compared to generative baselines, even when utilizing the LLM acceleration methods (see Appendix~\ref{appendix:llm_acceleration}).

\subsection{Ablation Study}
We evaluate the importance of the knowledge aggregation design in our model.
Our model utilizes an in-document phrase index by adding knowledge embedding from Wikipedia and phrase embedding from the input document.
In Table~\ref{tab:entity_linker}, (-External) excludes external knowledge embedding, and (-Internal) removes phrase embedding. 
Results indicate a notable performance drop with (-External) when both entity linkers are used.
When phrase embedding is removed (-Internal), the model with the Gold entity linker performs better overall, while the model with Wikifier shows lower results compared to using both embeddings. 
However, robustness of List Overlap scores consistently remains higher than when partial components are removed, emphasizing the vital role of internal knowledge in constructing a resilient embedding, particularly when external information quality is suboptimal.

%% file: KtrlF/8_user_study.tex
% User Study
\begin{figure}[!t]
\small
    \centering
    \subfloat[Number of queries]{
        \centering
        \includegraphics[width=0.23\textwidth]{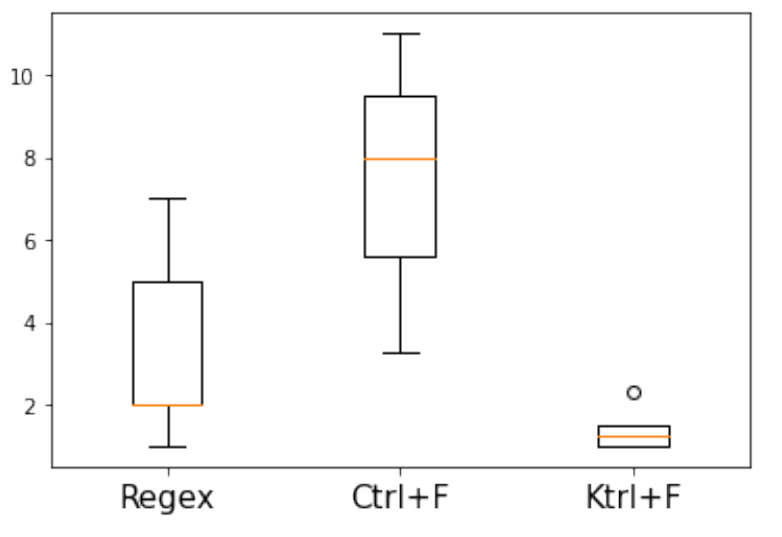}
        \label{fig:user_study_num_query}
    }
    % \hfill
    \subfloat[Number of visited websites]{
        \centering
        \includegraphics[width=0.23\textwidth]{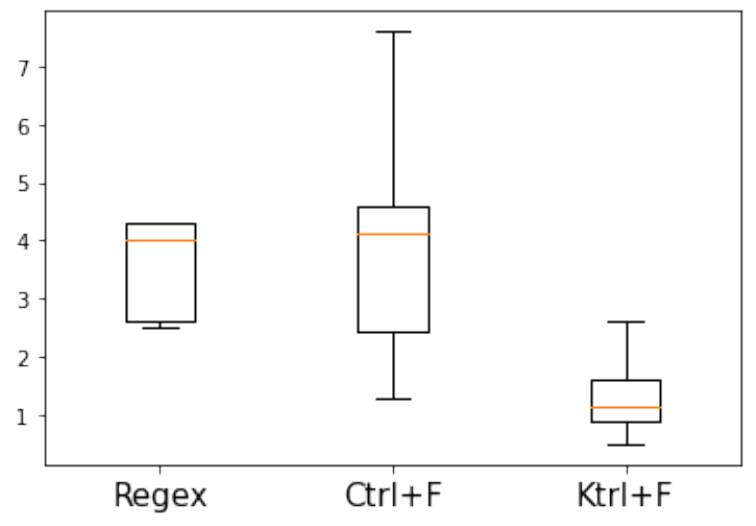}
        \label{fig:user_study_num_visit}
    }
    \hfill
    % \vspace{0.5cm}
    \subfloat[Spent time (sec)]{
        \centering
        \includegraphics[width=0.23\textwidth]{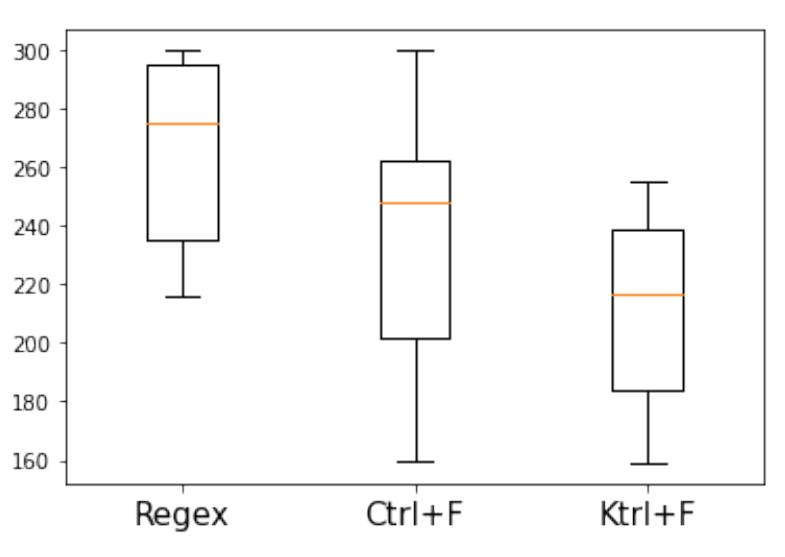}
        \label{fig:user_study_time}
    }
    % \hfill
    \subfloat[List EM F1 score]{
        \centering
        \includegraphics[width=0.23\textwidth]{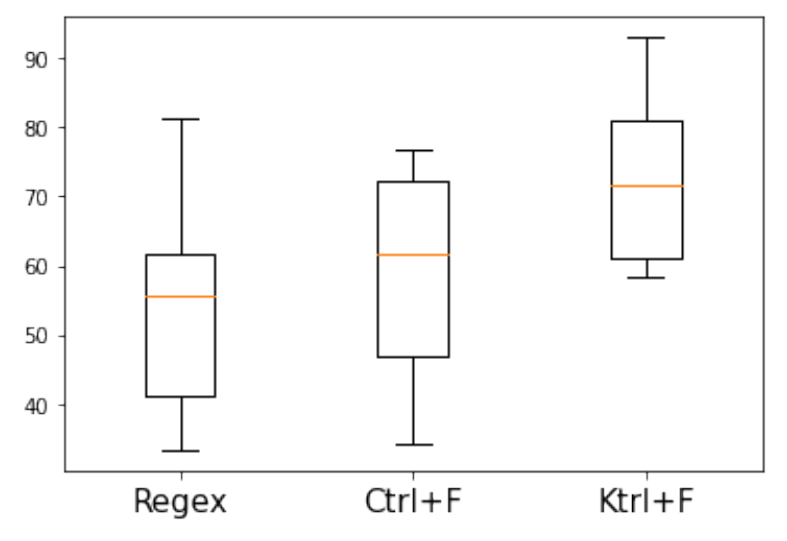}
        \label{fig:user_study_score}
    }
    \caption{
    A comparison of in-document search systems. Ktrl+F plugin outperforms other systems overall.
    }
    \label{fig:user_study}
    \small
% \vspace{-1em}
\end{figure}

% \vspace{-0.5em}
To verify whether solving \task can enhance search experience of users in the real web environments, 
we build Chrome extension plugin (\task plugin) built on our model.

\subsection{Setup}
% We recruit 6 participants for an in-depth experiment, where they are tasked with using the given system to find information from 10 different examples on a given website within a limited time~\footnote{We plan to expand more examples in future versions}.
% To ensure a fair comparison, each user is assigned, on an example-by-example basis, to use only a specific system among \task plugin, \ctrlf, and Regular expression to help them find all parts of targets that match given search intent.
% The participants manually annotate the targets in the PDFs using the respective system. To capture the entire process, we record the screens of participants throughout the experiment. During the analysis phase, we focus on several key elements. These include the time each participant spent finding information using the specified system, the number of queries used to find the target parts, the count of additional websites visited to verify the identified targets for each example, and an overall score for the prediction results using List EM F1 to evaluate the effectiveness of searching for multiple targets. 
% % For more detailed information, please refer to Appendix~\ref{appendix:user_study}.

% Update - final revision
% We assign a total of six individuals to solve 10 problems each. 
% For each example, we assigned two individuals per tool to enable us to collect responses using three different tools (Ctrl+F, Regex, KTRL+F plugin) for each example. 
% Instructions used are follow:

Each user is assigned to use only a specific system per example among \task plugin, \ctrlf, and Regular expression to help them find all targets that match given search intent from a given website.
Criteria for evaluation are shown in Figure~\ref{fig:user_study}.
% Add
Further details for the user study are provided in Appendix~\ref{appendix:user_study}.
% \vspace{-0.5em}

\subsection{Findings}
For a comprehensive comparison of the usefulness and efficiency of the \task plugin with other in-document search systems, we present the results of the conducted user study in Figure~\ref{fig:user_study}. 
% \vspace{-0.5em}
\paragraph{Less search time with \task plugin.}
As depicted in Figure~\ref{fig:user_study} \subref{fig:user_study_time}, the \task plugin exhibits the shortest time when searching for targets. 
This efficiency stems from its capacity to identify multiple semantic targets in a single query, minimizing the need for additional searches to validate results. 
While regular expressions can similarly search for multiple targets simultaneously, the process involves complex creation and often difficult debugging, as exemplified in Figure~\ref{tab:evaluation_query} of Appendix~\ref{appendix:user_study}.
% \vspace{-0.5em}

\paragraph{Fewer queries to find targets.}
Figure~\ref{fig:user_study} \subref{fig:user_study_num_query} illustrates the average number of queries used to find answers. 
Regular expressions and Ctrl+F rely on user-generated candidate lexical prefixes to find answers. 
Transforming search intent into the format supported by these systems increases query usage.
While \ctrlf allows swift query verification, users struggle to predict which keywords will appear in unknown text before reading it entirely.
Regular expressions can consolidate multiple simple searches into one, but dynamically crafting complex expressions is challenging and debugging erroneous code compounds the complexity.
% \vspace{-0.5em}

\paragraph{Fewer visits for extra sources.}
The ability to extend external knowledge beyond the current web page of \task plugin alleviates the need to consult additional sources to verify results, as shown in Figure~\ref{fig:user_study} \subref{fig:user_study_num_visit}.
Additionally, users often overlook variations when using manual lexical matching systems. 
For example, in the query "List all football teams from the web page," users might overlook variations such as Liverpool FC's nickname "The Reds." 
The ability to handle such subtle changes of \task plugin contributes to improved performance as  Figure~\ref{fig:user_study} \subref{fig:user_study_score}.

%% file: KtrlF/9_conclusion.tex
% \vspace{-0.5em}
In this paper, we introduce \task, a \TASK that requires identifying all semantic targets with a single natural query in real-time.  
\task tackles unique challenge for in-document search that requires capturing targets containing additional information beyond the input document by utilizing external knowledge while balancing speed and performance.
We highlight limitations in existing models, such as hallucinations, high latency, or difficulties to incorporate external knowledge.
And show that our \OURS, simple extension of phrase retrieval architecture can be a robust model for \task. 
Moreover, the study demonstrates that even our straightforward model, with seamless access to external knowledge during in-document searches, significantly enhances the user search experience.

Future work could extend \task to reflect updated knowledge, such as news, or domain-specific knowledge bases, such as the medical domain, which cannot be easily handled by large language models alone~\citep{ram2023context,peng2023check,kaddour2023challenges}.
The scalability and practicality of \task will open up opportunities for various advancements in the field of information retrieval and knowledge augmentation.

%% file: KtrlF/a_limitations.tex
% (2) Algorithmic contribution to the problem is of limited novelty (Reject #2)
% (4) Discussion on indexing stage of Ktrl+F and possible limitation (Question)
% developing a Knowledge-Augmented Ctrl+F plugin, with a primary focus on entities as search targets. 

The system design for \task can incorporate various forms of external knowledge, not limited to the Wikipedia page associated with the entity. It can also identify a wide range of target spans within the target text, including dates and numbers, without being restricted to entities.
However, the primary focus of this paper revolves around addressing \task, specifically emphasizing entities as the primary search targets. By narrowing our focus to entities, we make effective use of entity linking information as external knowledge. 
Furthermore, due to the inherent nature of retrieval systems, our \OURS model requires an extra indexing stage whenever a change in the input document, which requires additional time to use.
Also it relies on thresholding to truncate predicted results, which we employ top-k results based on the data distribution in our experiment.
Exploring more efficient methods for enhancing external knowledge while reducing the time needed for the indexing stage is a potential avenue.

%% file: KtrlF/a_acknowledgments.tex
This work was partly supported by Samsung Research grant (2021, Multi-grained Passage Embedding via Cross-to-Bi Encoder Distillation, 80\%) and Institute of Information \& communications Technology Planning \& Evaluation (IITP) grant funded by the Korea government (MSIT) (No.2021-0-02068, Artificial Intelligence Innovation Hub, 20\%).

%% file: KtrlF/a_appendix.tex
\section{Details for Dataset Construction Pipeline}
\label{appendix:dataset_details}
\paragraph{Step 1. Select Real News Articles.}
The preprocessing of articles involves two criteria.
First, 6,936 articles are collected from the 13,863 articles in the C4 realnewslike validation set, with lengths ranging from 991 to 3,298, covering the lower to upper quartiles to remove abnormal articles.
Then, to ensure diversity of questions and quality of documents, we collect 3,910 articles with 4 to 11 entities, covering the lower to upper quartiles.

We consider Wikipedia through October 31, 2023 as an external knowledge source. The acquisition of external knowledge for targets is equated to utilizing the corresponding Wiki page linked to a particular entity.~\citep{Wu2019ScalableZE}.

\paragraph{Step 3.1. Target Filtering.}
In this step, given a (query, entity, external knowledge) triple, we follow \citet{zhong2023romqa} to derive whether an entity is an answer to a query or not.
We utilize the first 10 sentences from the Wikipedia article as an external knowledge, which covers more than 99\% of the total sample within 4,096 tokens of GPT-3.5.
GPT-3.5 processes a total of 7,060 triple samples, and the final judgment is made by GPT-4 on 1,226 samples that show different results from the target generated by LLAMA-2 in Step 2.
On average, 1.6 entities disagreed per query, which is an average of 22\% of the candidate entities per query.
After the final judgment, queries with all targets determined to be false are discarded.
As a result, 816 queries remained out of the total 1,000 queries generated by Step 2, and the average number of entities in a target increased slightly from 1.4 to 1.9.

\paragraph{Step 3.2. Query Filtering.}
In this step, we exclude a query if the MRC model answers any of the target entities. 
The MRC model is considered correct when it scores over 0.9 in F1 score, following the human performance described in \citet{rajpurkar-etal-2018-know}.
As a result, 512 queries were collected from the 816 queries derived in Step 3-1.

\section{Inter-Annotator Reliability of Human Verification}
\label{appendix:inter-rater_reliability}
Eight annotators, all of whom are computer science majors proficient in English participated Human verification.
To assess the inter-annotator reliability among the three annotators, we utilize Fleiss' kappa value~\citep{Fleiss1971MeasuringNS}, a metric used to evaluate the agreement between multiple annotators in assigning categorical ratings. We follow the interpretation of kappa value by \citet{Landis1977TheMO}: < 0 indicates \textit{poor aggreement}; 0.01-0.20 indicates \textit{slight agreement}; 0.21–0.40 indicates \textit{fair agreement}; 0.41–0.60 indicates \textit{moderate agreement}; 0.61–0.80 indicates \textit{substantial agreement}; and 0.81–1.00 indicates \textit{almost perfect agreement}.

The first and second questions, classified as query-side verifications, scored kappa values of 0.552 and 0.4458 respectively, indicating \textit{moderate agreement} among the three annotators. In contrast, the third question scored  0.7193, indicating \textit{substantial agreement}. The nature of query-side verification, which relies on subjective evaluations, tends to result in lower inter-annotator reliability compared to target-side verification. The latter involves objective fact-checking with reference to Wikipedia, leading to higher agreement among annotators.

\section{Implementation Details for Baselines}
\paragraph{Generative baselines.}
\label{appendix:generative_methods}
To convert \task as generation problem, we use following instructions for generative models and then post-process the output text to only utilize the answer part. We use temparture 0.5, max new token 512.

\begin{lstlisting}[style=exp_prompt]
Find all mentions from the article 
below that correspond to the query. 
Only generate mentions with comma 
separate.

Article: {Input Document}
Query: {Query}
Mentions:  
\end{lstlisting}

\paragraph{Extractive baseline.}
% SequenceTagger 
We solve \task using sequence tagging model following \citep{li2022multispanqa}.
It can be regarded as a model without utilizing external knowledge.
We reproduce the model trained on MultiSpanQA~\citep{li2022multispanqa} for 3 epochs.

% UPDATE - final version
\section{Analysis of RAG baselines}
\label{appendix:rag_baseline}
% \paragraph{RAG baselines.}
We append the top-5 retrieved passages using DensePhrases~\citep{lee2021learning} as a retriever to the LLM input. Here is the prompt we utilized for the RAG experiment.

\begin{lstlisting}[style=exp_prompt]
Find all references to your query 
in the ARTICLE below, referring to 
the external evidence provided.

- Generates only matching pairs of 
  mentions from the ARTICLE, separated 
  by commas. Just generate answers! 
  This is IMPORTANT.
- Do NOT extract mentions from the 
  EVIDENCE.
- If a same mention appears multiple 
  time, generate every mentions. 
- Please do not generate any other 
  opening, closing, and explanations. 
  Just generate the set of scenarios!

#Evidence: {top-k paragraphs}
#Article: {target_text}
#Query:{query}
#Mentions: 
\end{lstlisting}

Despite explicitly providing additional information, incorporating retrieval information into the LLM input diminishes performance compared to a straightforward LLM approach. 
Notably, performance declines significantly across all models except for GPT-4, as demonstrated in Table~\ref{tab:appendix-rag}. 
Upon manual analysis, we observe that the retrieval system adequately retrieves paragraphs related to the query in general. 
However, two types of errors are identified: 1) failure to retrieve relevant targets for the target text during the retrieval stage, and 2) failure to ground instructions that not only extract information solely from the target text (the article) but also extract answers from the retrieved evidence during the generation stage.

% As an example, when the query is "Locations that have been designated as UNESCO World Heritage Sites," and the answer provided by the target is "Auschwitz," the RAG-GPT-4 model predicts an answer based on the evidence as follows:

% \begin{lstlisting}[style=exp_prompt]
% RAG PRED: ['Cartagena and its historic surroundings', 'the insular department of San Andrés', ' Providencia y Santa Catalina', 'Santa Marta and the surrounding area', 'Vigan', 'Santa Maria Church', 'Paoay Church', ' San Agustin Church', 'Miagao Church', 'Rice Terraces of the Cordilleras', 'Tubbataha Reefs', 'Underground River of Puerto Princesa', 'Mount Hamiguitan']
% \end{lstlisting}

For instance, when using 'Social media platforms' as the retrieval query, one of the retrieval results includes descriptions about various platforms such as Facebook, MySpace, YouTube, and blogs. 
However, in the corresponding target text to be skimmed, there are no relevant sections within the retrieved paragraphs, and the only target we can match from the provided target text is 'Twitter'.
In this scenario, the retrieved paragraphs can serve as distractors for the generative model, making it challenging to extract information solely from the target text, as indicated in the experimental table. 
We emphasize that the unique characteristics of our datasets in KTRL+F demand grounding information from both the query-side and target text side, presenting a distinct challenge. 

\begin{table*}[t]
\centering
\resizebox{1.\textwidth}{!}{
    \begin{tabular}{cccccccc}
    \toprule
    Model & List EM ($\uparrow$) & (R) List EM ($\uparrow$) & List Overlap ($\uparrow$)& (R) List Overlap ($\uparrow$) \\
     \midrule
     RAG-GPT-3.5 & 8.338 & 2.233 & 27.404 & 13.573\\
     RAG-GPT-4& \textbf{28.279} & \textbf{8.457} & \textbf{42.791} & \textbf{20.646} \\
     RAG-LLAMA-2-Chat-7B & 7.987 & 2.361 & 28.465 & 16.469\\
     RAG-LLAMA-2-Chat-13B &  9.140 & 2.262 & 26.949 & 12.894\\
     RAG-VICUNA-7B-v1.5 & 4.468	& 0.770 & 24.685 & 12.156\\
     RAG-VICUNA-13B-v1.5 & 5.773 & 1.163 & 28.745	& 16.860\\
     \bottomrule
    \end{tabular}
}
\caption{Results for RAG baselines. We utilize DensePhrases as a retriever and augment top 5 retrieved passages from the Wikipedia dump provided by the authors to the LLM input.
}
\label{tab:appendix-rag}
\end{table*}

\section{Further Analysis of Retrieval Approach for \task}
\label{appendix:metrics}

Determining a proper threshold for retrieval is challenging, especially when the number of targets varies.
Therefore, we additionally measure the Mean Average Precision (MAP), which calculates the mean value per query Q of the Area under the Curve (AUC) of the precision-recall graph in Table \ref{tab:appendix-map}.  
This metric provides a comprehensive measure of the system's ability to quantify the overall effectiveness.

\section{Prediction Examples}
\label{appendix:prediction_example}
Table~\ref{table:appendix_prediction} shows the results of various approaches on same query and input document for qualitative analysis.

\section{Baseline Analysis from Different Perspectives}
\label{appendix:additional_exeriments}
For a comprehensive baseline understanding, we additionally present set-base scores which doesn't require recognizing every target occurrences in Table \ref{tab:appendix-fair-comparison-set-score}. 
We can see the Set Overlap score gets a higher result than List Overlap overall, and especially generative models show major performance gain in Overlap score when using Set score, which shows finding all matching target is hard for generative models. 
Given that our model leverages entity linking information to identify targets from a restricted pool of candidates, we conduct an additional experiment by supplying additional information about potential targets for generative models (refer to Table~\ref{tab:appendix-fair-comparison}).
When adding extra information about potential targets for generative models, it proves to enhance the overall performance of generative models. 
Notably, in the case of LLM-API (GPT-4, GPT-3.5), it even outperforms our model with gold-standard information. 
However, it's important to note that enhancing information for generative models comes with increased costs and slower latency, making it impractical for real-time applicability.

\section{Details for User Study}
\label{appendix:user_study}
We compare existing in-document search systems in Table~\ref{tab:evaluation}, considering criteria such as matching type, the system's ability to search multiple targets, its search intention, and its capacity to augment external knowledge. 
Additionally, Table~\ref{tab:evaluation_query} includes examples of queries users employ with different in-document search systems to find the same targets.
 
% UPDATE - final revision
We recruit six participants from the computer science field, each solving 10 examples from designated websites.
For each example, we assign two individuals per tool to enable us to collect responses using three different tools (Ctrl+F, Regex, KTRL+F plugin) for each example. 
To present users with challenging search goals that require identifying multiple target variants within a document, we believe that leveraging the dedicated KTRL+F dataset tailored for this purpose was a natural choice. 
Thus, we select all examples linked to our Ktrl+F dataset.
The participants manually annotate the targets in the PDFs using the respective system. 
For in-depth analysis, all experiments are conducted on-site and we record the screens of participants throughout the experiment to capture the entire search process. 
Instructions given to the participants are as follows:

\begin{lstlisting}[style=exp_prompt]
- Click Web Page URL
- Find all candidate spans in the 
  Web page which meets noted search 
  intention.
- You can utilize Answer information 
  when you are using Ctrl+F & Regex
- NOTE: use only specified used system 
  per example 
- All extraction should be highlighted 
  manually in the linked PDF URL 
- You only have to fill spent time per 
  example manually in this sheet 
  (max 5min per example)

(FYI, You can search multiple targets 
 using Regex in this format : 
 \b(?:SAN JOSE|Calif|Anaheim)\b)
\end{lstlisting}

\section{Details for List Overlap F1 Metric}
\label{appendix:list_overlap_metric}
The List Overlap F1 score follows the definition of span overlap as outlined in MultispanQA~\cite{li2022multispanqa}. Equation~\ref{equation:list_overlap} calculates the partial retrieved and relevant scores for each pair ($p_i$, $g_i$) by determining the length of the longest common substring (LCS) and dividing it by the length of the respective spans.
% (Equation~\ref{equation:list_overlap},~\ref{equation:precision_recall_f1})
\begin{equation}
\begin{split}
s_{ij}^{ret} = len(LCS(p_i,g_i))/len(p_i) \\
s_{ij}^{rel} = len(LCS(p_i,g_i))/len(g_i)
\end{split}
\label{equation:list_overlap}
\end{equation}

Different from set-based F1, List Overlap identify all occurrences. When there are $n$ predicted occurrences and $m$ target occurrences for a question, all metrics are defined as below.

\begin{equation}
\begin{split}
Precision & = \frac{1}{n} \displaystyle \sum_{i=1}^{n} \max_{j\in [1,m]} (s_{ij}^{ret}) \\
Recall & = \frac{1}{m} \displaystyle \sum_{j=1}^{m} \max_{i\in [1,n]} (s_{ij}^{rel}) \\
F1 & = \frac{2 \times Precision \times Recall}{Precision + Recall}
\end{split}
\label{equation:precision_recall_f1}
\end{equation}

\section{Comparison with LLM Acceleration Methods}
\label{appendix:llm_acceleration}
In the Table~\ref{tab:main_result}, the generative baselines show poor latency relative to its performance. We compare how much the generative method can compensate for latency through acceleration methods, including algorithmic acceleration methods such as Lookahead Decoding~\cite{fu2024break} and Medusa~\cite{cai2024medusa}, as well as hardware-level acceleration such as vLLM~\cite{kwon2023efficient}. The Medusa~\cite{cai2024medusa} shows nearly 2x speedup, but still lagging behind retrieval methods. However, even without any low-level optimizations, our retrieval-based method is still more efficient than generative approaches. Considering real-time latency as a key requirement for KTRL+F, exploring generative approaches in this problem holds promise for future research.

\begin{table}[h!]
% \begin{minipage}{\columnwidth}
\centering
\resizebox{0.3\textwidth}{!}{
    \begin{tabular}{lcccc}
    \toprule
     Model & Latency (ms/Q) ($\downarrow$) \\
     \midrule
     Vicuna-7B-v1.5 & 1951 \\
     + Lookahead & 1520 \\
     + vLLM & 1277 \\
     + Medusa & 1012 \\
    \midrule
     Vicuna-13B-v1.5 & 2420 \\
     + Lookahead & 2046 \\
     + vLLM & 1749 \\
     + Medusa & 1280 \\
    \midrule
    Ours & \textbf{15} \\
     \bottomrule
    \end{tabular}
}
\caption{Comparison of latency on Vicuna with acceleration methods.}
\label{tab:acceleration_methods}
\end{table}

\begin{figure*}[t]
\begin{lstlisting}[style=data_prompt]
You are an expert of query generation for entity search.

You must follow this requirements.
Requirements:
- Your task is to generate queries that retrieve entities in a given list.
- The generated query must be able to list the multiple entities.
- The answers must be countable.
- The answers have to be entities.
- Make sure your questions are unambiguous and based on facts rather than temporal information.
- Do not specify the number in a query.
- Do not start with 'What' in a query.
- Do not start with 'Which' in a query.
- Do not include an expression in your query that tells it to find from a given list.

The example is as below.

Generate 4 queries from the following list and extract subset list.
Candidate List:
[Apple, Microsoft, Samsung Electronics, Alphabet, AT&T, Amazon, Verizon Communications, China Mobile, Walt Disney, Facebook, Alibaba, Intel, Softbank, IBM, Tencent Holdings, Nippon Telegraph & Tel, Cisco Systems, Oracle, Deutsche Telkom, Taiwan Semiconductor, KDDi , HP, Legend Holding, Lenovo Group, ebay]

Query:
1. IT companies in Computer Hardware industry
=> Apple, HP, Legend Holding, Lenovo Group
2. Find all IT companies that have software as main business.
=> Microsoft, Oracle
3. Companies that is known for retail service
=> Amazon, Alibaba, ebay
4. Name all IT companies that have license in USA
=> Apple, Microsoft, Alphabet, AT&T, Amazon, Verizon Communications, Walt Disney, Facebook, Intel, IBM, Cisco Systems, Oracle, HP, ebay
\end{lstlisting}
\caption{Prompt for generating queries and targets}
\label{fig:query_generation_prompt}
\end{figure*}

\begin{figure*}[t]
\begin{lstlisting}[style=data_prompt]
You are a QA system to identify the given entity is the answer.
The inputs are entity, query and evidence.

You must follow this requirements.
Requirements:
- Output have to be either 'true' or 'false'
- Do not say anything except 'true' or 'false'

The example is as below.

Entity: Google
Query: Find all IT companies in Computer industry
Evidence: Google LLC is an American multinational technology company focusing on artificial intelligence, online advertising, search engine technology, cloud computing, computer software, quantum computing, e-commerce, and consumer electronics. It has often been considered "the most powerful company in the world" and as one of the world's most valuable brands due to its market dominance, data collection, and technological advantages in the field of artificial intelligence. Its parent company Alphabet is often considered one of the Big Five American information technology companies, alongside Amazon, Apple, Meta, and Microsoft.
Output: true

Entity: Samsung
Query: Find all companies in United States
Evidence: Samsung Group, or simply Samsung, is a South Korean multinational manufacturing conglomerate headquartered in Samsung Town, Seoul, South Korea. It comprises numerous affiliated businesses, most of them united under the Samsung brand, and is the largest South Korean chaebol (business conglomerate). As of 2020, Samsung has the eighth highest global brand value.
Output: false
\end{lstlisting}
\caption{Prompt for target filtering}
\label{fig:answer_refine_prompt}
\end{figure*}

\begin{figure*}[t]
    \centering
    \subfloat[The Q1 requests the identification of evidence for each target to evaluate whether the query satisfies \reqtw.]{
        \centering
        \includegraphics[width=0.7\textwidth]{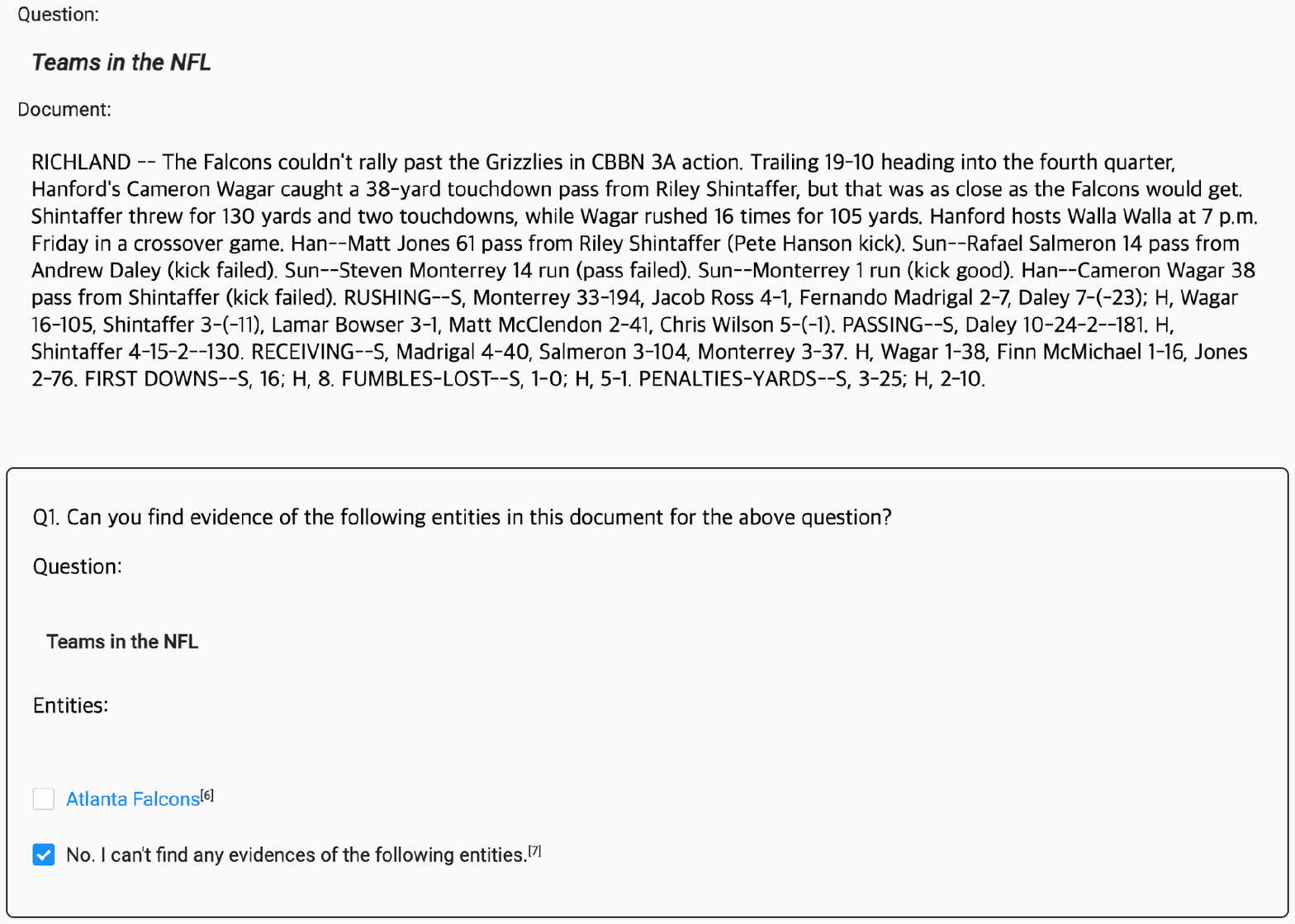}
    } \\
    \subfloat[The Q2 requests the selection of options to evaluate the naturalness of the query.]{
        \includegraphics[width=0.7\textwidth]{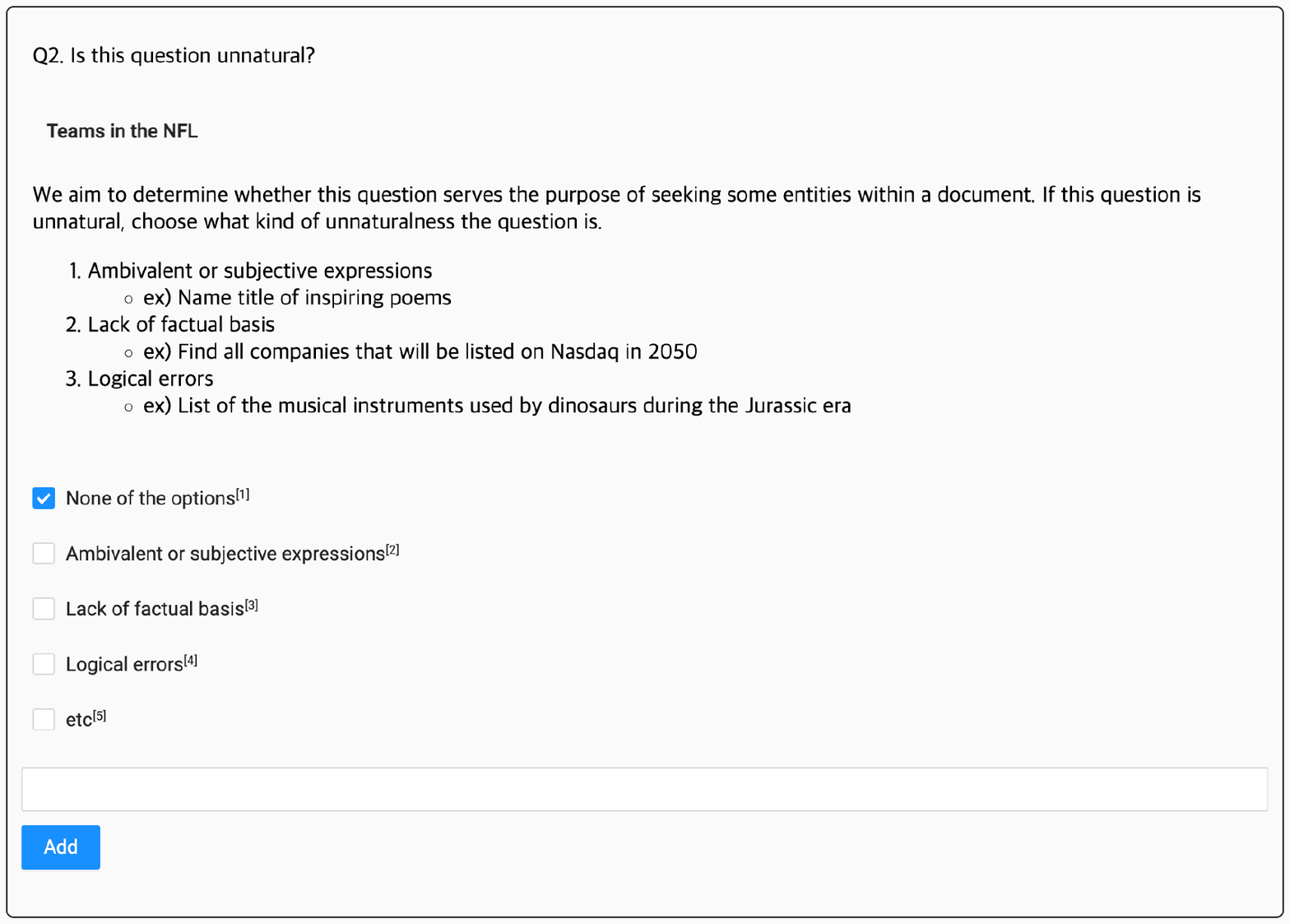}
    } \\
    \subfloat[The Q3 requests the selection of targets to evaluate the reliability of target determination.]{
        \includegraphics[width=0.7\textwidth]{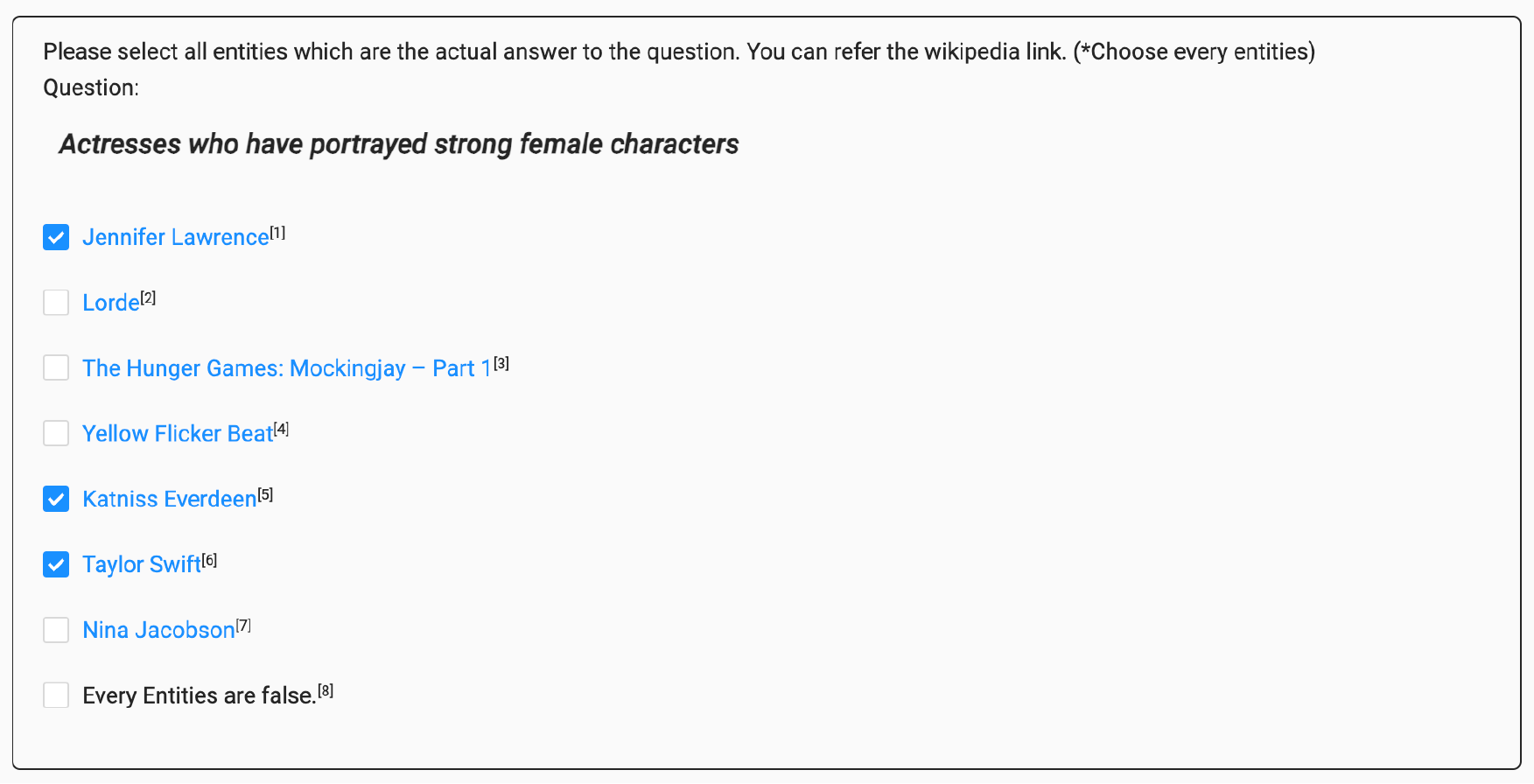}
    } 
    \caption{User Interface for Human Verification.}
    \label{fig:human_verification_page}
\end{figure*}

\begin{table*} [t]
\centering\
\resizebox{\textwidth}{!}{
\def\arraystretch{0.9}
\begin{tabularx}{\textwidth}{X}
\toprule
\small\textbf{[Query] \textit{Entities that are known for their cookie products}} \\
\small\textbf{[Input Document]} \\ 
\small \colorbox{yellow}{Nabisco} threatens to sue a Canadian man who registered "\colorbox{yellow}{oreos.com}" for his home page with adult links. "\colorbox{yellow}{Oreos.com}" sat quietly on the Net for more than a year--however, it wasn't a hub to debate whether the cookie's crunchy chocolate outside is better than its creamy filling. On the contrary, until today, "\colorbox{yellow}{Oreos.com}" was an Ontario man's personal Web page featuring links to some adult entertainment sites. While this may have been a treat for some, it is not exactly the one most people affiliate with \colorbox{yellow}{Nabisco}'s famous sandwich cookie. And so it was that Paul Figueiredo found himself in a legal dispute with one of the biggest food companies in the United States and Canada. \colorbox{yellow}{Nabisco} threatened a lawsuit if he didn't surrender the domain name by noon today. ... At first, "\colorbox{yellow}{Oreos.com}" was registered to be the site for the Ontario Real Estate Online Services, for advertising homes on the market, Figueiredo says. But his business idea never took off. He had already spent \$100 to register the site name, so he turned it into a home page. When he got the letter from \colorbox{yellow}{Nabisco}'s lawyers earlier this month, he knew the site's days were numbered. ... He tried unsuccessfully to cut a deal that would have allowed him to point people to \colorbox{yellow}{Nabisco}'s official site, if he took the adult links off the front page. But because the company markets its sweets to kids, \colorbox{yellow}{Nabisco} wouldn't go for it, he said. "If it was my kid, I wouldn't want them to see adult banners when they type in '\colorbox{yellow}{Oreos},'" he admitted. ... And most won't take "no" for an answer. "He has faxed back the letter agreeing to cease all use of the \colorbox{yellow}{Oreos} trademark and domain name," Jonathan Colombo, an attorney for \colorbox{yellow}{Nabisco}, said today. \\
\midrule
\small\textbf{[Query] \textit{Companies that offer cloud computing services}} \\
\small\textbf{[Input Document]} \\ 
\small  Razer's latest eGPU cabinet gets LEDs and a bigger PSU , plus a ton more ports than before. Alienware's redesigned powerhouse laptop promises the Holy Grail of gaming laptop features. It's big, fast, beautiful, and even upgradable. \colorbox{yellow}{Google}'s shown it can kill off a product when it no longer deserves to live. We know a few more products that are ready to die, if only \colorbox{yellow}{Google} could help. We go hands-on with HP's Reverb Consumer Edition, whose astounding resolution is well deserving of this exclamation mark! Here's what you need to know about Maxon's new Cinebench R20 benchmark, and how to use it to test your computer. Acer's Predator Helios 300 is currently the bestselling gaming laptop on \colorbox{yellow}{Amazon}. With an 8th-gen Core i7, GeForce GTX 1060, and 144Hz screen, it's easy to see why. We delve into those and other details. Give the ThinkPad six cores and a GeForce GPU and you get the Lenovo ThinkPad Extreme X1, a 15-inch laptop that's large and in charge. Lenovo's newest mainstream IdeaPad laptops give you a choice between Ryzen and RX Vega, or Core i7 and a mystery GeForce MX graphics. \\
\midrule
\small\textbf{[Query] \textit{Cities in Wisconsin}} \\
\small\textbf{[Input Document]} \\ 
\small Workers wear double-lined suits, and the floor is heated to prevent permafrost. In one of the coldest workplaces on earth, in \colorbox{yellow}{New Berlin}, employees wear heated boots with a 2-inch-thick sole. Inside their work area — two freezers totaling 12,000 square feet — it’s nearly 70 below zero, colder than most winter days in Siberia. ... Cultures are stored at minus-67 degrees until they’re shipped, frozen, to food companies that thaw them and put them to work making products. The company also makes probiotic bacteria strains for health care companies around the world. “We develop and produce cultures, enzymes, probiotics and natural colors for a rich variety of foods, confectionery, beverages, dietary supplements and even animal feed and plant protection,” the company says. More than 1 billion people a day consume products containing the company’s natural ingredients, the Chr Hansen website says. The company has more than 3,000 employees, in about 30 countries, including about 300 in \colorbox{yellow}{New Berlin} and the \colorbox{yellow}{Milwaukee} area. It was founded by a Danish pharmacist in 1874 and has been in the \colorbox{yellow}{Milwaukee} area since the late 1920s. “We’ve been pretty fortunate in the people we’ve been able to recruit and retain,” Graham said. \\
\midrule
\small\textbf{[Query] \textit{Sports teams in the state of Georgia}} \\
\small\textbf{[Input Document]} \\ 
\small RICHLAND -- \colorbox{yellow}{The Falcons} couldn't rally past the Grizzlies in CBBN 3A action. Trailing 19-10 heading into the fourth quarter, Hanford's Cameron Wagar caught a 38-yard touchdown pass from Riley Shintaffer, but that was as close as the \colorbox{yellow}{Falcons} would get. Shintaffer threw for 130 yards and two touchdowns, while Wagar rushed 16 times for 105 yards. Hanford hosts Walla Walla at 7 p.m. Friday in a crossover game. Han--Matt Jones 61 pass from Riley Shintaffer (Pete Hanson kick). Sun--Rafael Salmeron 14 pass from Andrew Daley (kick failed). Sun--Steven Monterrey 14 run (pass failed). Sun--Monterrey 1 run (kick good). Han--Cameron Wagar 38 pass from Shintaffer (kick failed). ... \\
\bottomrule
\end{tabularx}}
\caption{Example of \task evaluation dataset. The highlights indicate target mentions and link to the Wikipedia page. For example, in the fourth sample, "Falcons" links to the Wikipedia page for "Atlanta Falcons".}
\label{table:appendix_data_example}
\end{table*}

\begin{table*}[t!]
\begin{minipage}{\textwidth}
\centering
\resizebox{0.7\textwidth}{!}{
    \begin{tabular}{ccccccc}
    \toprule
     & \multicolumn{3}{c}{List EM} & \multicolumn{3}{c}{List Overlap} \\     
     \cmidrule(lr){2-4} 
     \cmidrule(lr){5-7}
    Model & Precision & Recall & F1 & Precision & Recall & F1 \\
     \midrule
     GPT-3.5 & 39.2 & 33.0 & 30.3 & 54.2 & 49.3 & 41.9 \\
     GPT-4 &  39.0 & 31.8 & 30.4 & 49.0 & 41.6 & 37.4 \\
     LLAMA-2-Chat-7B & 28.5 & 35.9 & 28.5 & 46.6 & 49.9 & 40.5 \\
     LLAMA-2-Chat-13B & 37.5 & 29.5 & 28.8 & 50.1 & 40.9 & 37.0 \\
     VICUNA-7B-v1.5 & 24.3 & 21.9 & 17.8 & 39.2 & 44.9 & 31.2 \\
     VICUNA-13B-v1.5 & 29.1 & 36.1 & 24.4 & 43.5 & 56.1 & 39.2\\
     \midrule
     SequenceTagger  & 12.6 & 6.3 & 7.23 & 18.8 & 7.6 & 8.6 \\
     \midrule
     Ours (w/ Wikifier)  & 23.7 & 33.5 & 23.1 & 48.3 & 47.4 & 40.7\\
     Ours (w/ Gold) & 47.7 & 63.6 & 46.1 & 61.2 & 64.6 & 53.6 \\
     \bottomrule
    \end{tabular}
}
\caption{Detailed performance evaluation including Precision and Recall for \task dataset.}
\label{tab:precision_recall}
\end{minipage}
\end{table*}

\begin{table*}[t]
% \begin{minipage}{\columnwidth}
\centering
\resizebox{1.0\textwidth}{!}{
    \begin{tabular}{ccccc}
    \toprule
     Model & Indexing time (Sec) ($\downarrow$) & ms/Q ($\downarrow$) & MAP(@IoU0.5) ($\uparrow$)&(R)MAP(@IoU0.5) ($\uparrow$) \\
     \midrule
     Ours w/ Wikifier & 3.555 & 14 & 0.464 & 0.209\\
     w/o INT & 3.027 & 14 & 0.494 & 0.220 \\
    w/o EXT & 3.145 & 14 & 0.335 & 0.153\\
    \midrule
     Ours w/ Gold & 0.955 & 14 & 0.716 & 0.380\\
     w/o INT & 0.912 & 14 &  0.776 & 0.408\\
     w/o EXT & 0.799 & 14 & 0.508 & 0.213\\
     \bottomrule
    \end{tabular}
}
\caption{MAP metric for retrieval approach. The result shows the effectiveness of phrase retrieval architecture. When using MAP as a metric, it reflect retrieved ranks of results and ours show slightly performance drop than ours w/o internal knowledge. }
\label{tab:appendix-map}
\end{table*}

\begin{table*}[t]
\centering
\resizebox{1.0\textwidth}{!}{
    \begin{tabular}{cccccccc}
    \toprule
    Model & List EM ($\uparrow$) & Set EM ($\uparrow$) & List Overlap ($\uparrow$)& Set Overlap ($\uparrow$) \\
     \midrule
     GPT-4 & 30.457 & 36.422 & 37.402 & 51.071 \\
     GPT-3.5 & 30.346 & 36.668 & 41.929 & 56.334 \\
     LLAMA-2-Chat-7B & 28.529 & 34.235 & 40.546 & 52.843\\
     LLAMA-2-Chat-13B  & 28.846 & 35.206  & 37.098 & 51.672\\
     VICUNA-7B-v1.5  & 17.831 & 22.265 & 31.216 & 42.460\\
     VICUNA-13B-v1.5 & 24.490 & 29.223 & 39.278 & 49.449\\
     \midrule
     SequenceTagger & 7.239 & 9.041 & 8.614  & 15.648 \\
     \midrule
     \OURS(w/ Wikifier) & 23.152 & 24.793 & 40.718  & 46.841 \\
     \OURS(w/ Gold) & \underline{46.170} & \underline{50.254} & \underline{53.689} & \underline{63.230} \\
     \bottomrule
    \end{tabular}
}
\caption{We additionally report Set-based scores with our List-based scores, which doesn't necessitate recognizing every target occurrences. 
}
\label{tab:appendix-fair-comparison-set-score}
\end{table*}
 
\begin{table*}[t]
\centering
\resizebox{1.\textwidth}{!}{
    \begin{tabular}{cccccccc}
    \toprule
    Model & List EM ($\uparrow$) & (R) List EM ($\uparrow$) & List Overlap ($\uparrow$)& (R) List Overlap ($\uparrow$) \\
     \midrule
    GPT-4 (w/ Gold)& \textbf{52.937} & \textbf{22.479} & \underline{55.765} & 25.183 \\
     GPT-3.5 (w/ Gold) & 44.697 & 22.048 & \textbf{56.615} & \textbf{35.874} \\
     LLAMA-2-Chat-7B (w/ Gold) & 40.225 & 17.738 & 50.466 & 30.140\\
     LLAMA-2-Chat-13B (w/ Gold) &  45.674 & 19.329 & 50.172 & 23.291\\
     VICUNA-7B-v1.5 (w/ Gold) & 27.374 & 8.651 & 41.466 & 21.611\\
     VICUNA-13B-v1.5 (w/ Gold) & 39.898 & 17.065 & 54.695 & 33.814 \\
     \midrule
     % \OURS(w/ Wikifier) & 23.152 & 24.793 & 40.718  & 46.841 \\
     \OURS(w/ Gold) & \underline{46.170} & \underline{22.426} & 53.689 & \underline{32.285} \\
     \bottomrule
    \end{tabular}
}
\caption{Results for when generative models use candidate entities from input document as additional input for instruction (denoted as w/ Gold). 
We evaluate the results by giving gold entity linking information version.
}
\label{tab:appendix-fair-comparison}
\end{table*}

\begin{table*} [t]
\centering\
\resizebox{\textwidth}{!}{
\def\arraystretch{0.9}
\begin{tabularx}{\textwidth}{X}
\toprule
\small\textbf{[Query] \textit{Social network platform of China}} \\
\small\textbf{[Input Document]} \\ 
\small It is a highly competitive market with many local competitors who already understand the shopping habits of the Chinese, which are very different to those of consumers in the Western world. Chinese platforms such as Taobao and Tmall dominate the shopping world …  successfully. \colorbox{yellow}{BAIDU} DOMINATES ONLINE Currently, there are an estimated 900 million internet users across China, with most users spending 1.5 hours a day just browsing. \colorbox{yellow}{Baidu} is the most popular search engine across China. Think of it as ‘the Google of China’. … time. \colorbox{yellow}{Baidu} also brings the … social media app across China, it is imperative that your company becomes familiar with the inner workings of \colorbox{yellow}{Wechat}. \colorbox{yellow}{Wechat} is fundamentally a social media… in store through \colorbox{yellow}{Wechat} pay are all possible. Many local and foreign companies already successful in China have official accounts on \colorbox{yellow}{Wechat}. An official … \colorbox{yellow}{Wechat} wallet. Many companies now also offer customer service through \colorbox{yellow}{Wechat}. Again, this is highly advisable as this is a service many Chinese consumers will now look for as it is quick and direct. \colorbox{yellow}{Weibo} is another popular social media \colorbox{yellow}{app} used across China. Think of \colorbox{yellow}{Weibo} as ‘the Twitter of China’. \colorbox{yellow}{Weibo} is an open network site so users can see posts from anyone without being their friend or following them. Similar to Twitter, \colorbox{yellow}{Weibo} can be an excellent way to market your company by sharing the latest updates, offers, promotions etc. Your followers can also start to share your content helping your company’s reputation spread by word of mouth. \\
\small\textbf{[Ground Truth]}
\small ['Wechat', 'Wechat', 'Wechat', 'Wechat', 'Wechat', 'Wechat', 'Weibo', 'network site', 'Weibo', 'Weibo', 'app', 'Weibo', 'Baidu', 'Baidu', 'BAIDU']\\
\midrule
\small\textbf{GPT-4}: ['Wechat', 'Weibo']  \\\midrule
\small\textbf{GPT-3.5}: ['Taobao', 'Tmall', 'Wechat', 'Weibo'] \\\midrule
% \small\textbf{Vicuna-7B}: ['WeChat'
% 'Sina Weibo', 'China'] \\\midrule
\small\textbf{Vicuna-13B-v1.5}: ['WeChat (official accounts, WeChat pay, customer service)', 'Taobao (dominates shopping world)', 'Tmall (dominates shopping world)', 'Baidu (most popular search engine, brings WeChat)','Weibo (open network, excellent way to market, Twitter of China)']\\\midrule
\small\textbf{SequenceTagger}: ['taobao, 'tmall'] \\\midrule
\small\textbf{Ours (w/o INT)}: ['service', 'Wechat', 'Wechat', 'Wechat', 'Wechat', 'Wechat', 'Wechat', 'way', 'Weibo', 'network site', 'Weibo', 'Weibo', 'app', 'Weibo', 'Taobao', 'Tmall', 'Twitter', 'Twitter', 'China', 'China', 'China', 'China', 'China', 'China', 'China', 'Baidu', 'Baidu', 'BAIDU', 'Chinese', 'Chinese', 'Chinese']\\\midrule
\small\textbf{Ours (w/o EXT)}: ['Weibo', 'Weibo', 'Wechat', 'Wechat', 'Weibo', 'BAIDU', 'Weibo', 'Baidu', 'Wechat', 'Twitter', 'Taobao', 'Tmall', 'Wechat', 'Wechat', 'Baidu', 'China', 'Wechat', 'China', 'app', 'Twitter', 'China', 'China', 'Chinese', 'network site', 'China', 'Chinese', 'China', 'Chinese', 'China', 'way', 'service']\\\midrule
\small\textbf{Ours}: ['Wechat', 'Wechat', 'Weibo', 'Weibo', 'Wechat', 'Wechat', 'Weibo', 'Wechat', 'Weibo', 'Wechat', 'Taobao', 'app', 'network site', 'Tmall', 'Twitter', 'BAIDU', 'Baidu', 'service', 'China', 'China', 'Twitter', 'Baidu', 'China', 'way', 'China', 'China', 'China', 'China', 'Chinese', 'Chinese', 'Chinese']\\
\bottomrule
\end{tabularx}}
\caption{Prediction result per different approaches. Note that our model uses thresholding for find proper points per query. In this result we show all ranking results.}
\label{table:appendix_prediction}
\end{table*}

\begin{table*}[t!]
\small
\resizebox{\textwidth}{!}{%
\begin{tabular}{c | ccccc}
\toprule
 & Matching Type & Search Mulitple Targets & Search Intention &  External Knowledge-Augmented  \\
\midrule
\midrule
Ctrl+F & Lexical & NO & Skimming & Manual &  \\
\midrule
Regular Expression & Lexical & YES  & Skimming &  Manual &  \\
\midrule
MRC & Semantic & YES & After Understanding &  NO &  \\
\midrule
\textbf{\task} & Semantic & YES & Skimming &  Automatic& \\
\bottomrule
\end{tabular}%
}
\caption{Comparing characteristics of \task with other systems.}
\label{tab:evaluation}
\end{table*}

\begin{table*}[!]
\small
\resizebox{\textwidth}{!}{%
\begin{tabular}{c|ccccc}
\toprule
 & Time(s) & \# of Queries  & \# of visited Websites &  Performance(List EM F1) \\
\midrule
Ctrl+F & 235(248) & 7.47(8) & 3.95(4.12) & 58.64(61.79)\\
\midrule
Regular Expression & 265(275) & 3.4(2) & 3.54(4) & 54.31(55.74)\\
\midrule
\textbf{\task plugin} & 211(217) & 1.41(1.25) & 1.08(1) & 72.70(71.60)\\
\bottomrule
\end{tabular}%
}
\caption{Evaluation table for comparing \task plugin with other systems. Averaged value is reported and median value are noted within bracket.}
\label{tab:evaluation_detail}
\end{table*}

\begin{table*}[!]
\small
\begin{tabular}{m{4cm}|m{4cm}|m{8cm}}
 \toprule
 Search Intention & Query per System & Result \\
     \midrule
      List the cities from California & \textbf{Ktrl+F :}  \small List the cities from California &   \small \colorbox{yellow}{SAN JOSE}, \colorbox{yellow}{Calif}. - Paramount to the ... they played smarter than they did Sunday in \colorbox{yellow}{Anaheim}, ... The Rangers signed 23-year-old defenseman Vince Pedrie out of Penn State, for whom he had 30 points in 39 games this season. \\
     \cmidrule{2-3}
      & \textbf{Ctrl+F :} \small [San jose, California, Anaheim] &   \small \colorbox{yellow}{SAN JOSE}, \colorbox{red}{Calif}. - Paramount to the ... they played smarter than they did Sunday in \colorbox{yellow}{Anaheim}, ... The Rangers signed 23-year-old defenseman Vince Pedrie out of Penn State, for whom he had 30 points in 39 games this season. \\
     \cmidrule{2-3}
     & \textbf{Regex: }\small(SAN JOSE | California | Anaheim) &  \small \colorbox{yellow}{SAN JOSE}, \colorbox{red}{Calif}. - Paramount to the ... they played smarter than they did Sunday in \colorbox{yellow}{Anaheim}, ... The Rangers signed 23-year-old defenseman Vince Pedrie out of Penn State, for whom he had 30 points in 39 games this season. \\
     \midrule
     List all football teams &  \textbf{Ktrl+F :} \small List all football teams &  \small \colorbox{yellow}{LIVERPOOL} star Fabinho has been caught on camera appearing to sneeze on \colorbox{yellow}{Chelsea}'s Eden Hazard. \colorbox{yellow}{Liverpool} took back top spot in the Premier League after beating \colorbox{yellow}{Chelsea} at Anfield earlier today. The \colorbox{yellow}{Reds} now have four games ... leading \colorbox{yellow}{Manchester City} by ... “He’s a fantastic player. \colorbox{yellow}{Chelsea} is ...\\
     \cmidrule{2-3}
      &  \textbf{Ctrl+F :} \small [Liverpool, Chelsea, Manchester City] &  \small \colorbox{yellow}{LIVERPOOL} star Fabinho has been caught on camera appearing to sneeze on \colorbox{yellow}{Chelsea}'s Eden Hazard. \colorbox{yellow}{Liverpool} took back top spot in the Premier League after beating \colorbox{yellow}{Chelsea} at Anfield earlier today. The \colorbox{red}{Reds} now have four games ... leading \colorbox{yellow}{Manchester City} by ... “He’s a fantastic player. \colorbox{yellow}{Chelsea} is ...\\
     \cmidrule{2-3}
     & \textbf{Regex:} \small (LIVERPOOL | Chelsea | Manchester City) &  \small \colorbox{yellow}{LIVERPOOL} star Fabinho has been caught on camera appearing to sneeze on \colorbox{yellow}{Chelsea}'s Eden Hazard. \colorbox{yellow}{Liverpool} took back top spot in the Premier League after beating \colorbox{yellow}{Chelsea} at Anfield earlier today. The \colorbox{red}{Reds} now have four games ... leading \colorbox{yellow}{Manchester City} by ... “He’s a fantastic player. \colorbox{yellow}{Chelsea} is ...\\
 \bottomrule
\end{tabular}
\caption{The figure above illustrates how each system handles the same search intention. It is worth noting that Ctrl+F and Regex require additional search engines to convert natural language search intentions, such as \textit{"List the cities from California,"} into candidate keywords like \textit{"Los Angeles, San Diego, San Jose, San Francisco, etc."} which consist of over a thousand cities. Moreover, there is no guarantee that these cities will appear on the web page. The highlighted text in yellow represents potential correct targets based on the query, while the red indicates possible false negative failures when using lexical search systems like Ctrl+F and Regex, which need to be highlighted.}
\label{tab:evaluation_query}
\end{table*}

\begin{figure*}[p!]
\small
\centering
\includegraphics[width=0.7\textwidth]{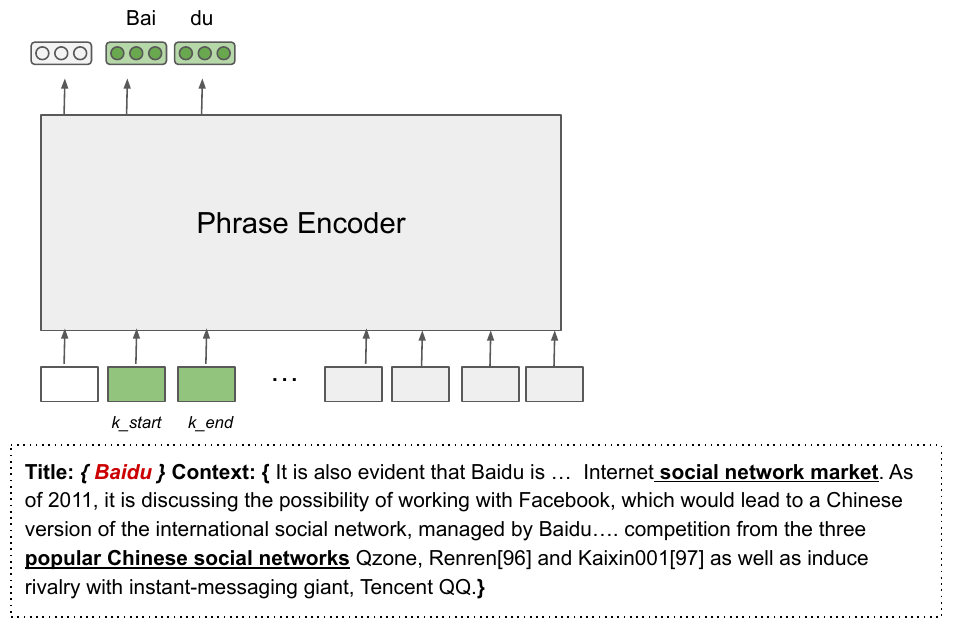}
\caption{The figure demonstrates how to extract knowledge embedding, which is used for external knowledge for \OURS.
We utilize the frozen pre-trained phrase retrieval model \citep{lee2021learning}, which shows good at encoding contextual information. The idea of using concatenated text with title and context and only extracting title embedding are following \citep{Lee2022ContextualizedGR}
}
\label{fig: method_external_embedding}
\end{figure*}

\begin{table*} [t]
\centering\
\resizebox{\textwidth}{!}{
\def\arraystretch{0.9}
\begin{tabularx}{\textwidth}{X}
\toprule
\small\textbf{[Query] \textit{Companies founded by Bill Gates}} \\
\small\textbf{[Input Document]} \\ 
\small That’s a line remote co-workers often ask each other when they need to really discuss something, face-to-face or at least orally. But later this year, both of those software programs could find themselves sidelined by Slack.
The makers behind the chat app announced yesterday that Slack users who are messaging each other will soon be able to have a voice call as well, and eventually a video call. No timeline has been given for either feature.
Slack’s rapid rise has already made it a darling of Silicon Valley. Just a year after its launch, investors valued the business chat app at over \$1 billion. It was pegged at \$2.8 billion as of last April, despite annual recurring revenues of just \$25 million. That valuation is thanks to Slack’s fast-growing, devoted customer base, which has skyrocketed from 500,000 daily active users in January 2015 to 2.3 million daily active users today.
What percentage of these users are also Skype or Google Hangout users is impossible to say, but judging by purely anecdotal evidence from people in the tech and new media world, there’s huge crossover. Most people who use Slack for business use a combination of Skype and Google Hangouts when they need to talk to someone face-to-face. But a majority of their time remains inside Slack, where they can write text messages to individual colleagues as well as set up team channels, upload gifs, and use special tools.
Google is likely indifferent to Slack’s rise. It makes almost all of its money from advertisements, and options like Google Hangouts are just there to keep users close to the search bar.
But Skype is a different story.
Skype, which is part of \colorbox{yellow}{Microsoft}, is mostly a two-trick app, and used for voice and video calls. Skype probably has more users that Slack right now, but some of its most valuable, paying “business” users are likely to fall into that “also using Slack” category. The company claimed 300 million monthly active users as of date, and had 4.9 million active daily users at the end of 2014, the latest daily data available.
Skype for Business, the app’s business-centric upgrade, is only a hair more sophisticated than regular Skype, which at the ripe old age of 12 both looks and feels dated. It has none of the bells and whistles that make Slack addictive in the first place. Adding voice and video calling to Slack could help businesses use one tool for nearly all communication purposes.
Skype is owned by \colorbox{yellow}{Microsoft}, and accounts for a tiny percentage of the \$93 billion in sales \colorbox{yellow}{Microsoft} brought in last year (they aren’t broken out, but in 2013 \colorbox{yellow}{Microsoft} told Bloomberg Skype was set to bring in \$2 billion in annual sales). \colorbox{yellow}{Microsoft} has been focusing on Windows Azure, an Amazon Web Services competitor, and revamping Skype is probably not a major priority. \colorbox{yellow}{Microsoft} did not respond to Quartz’s requests for comment on this piece.
But how many young companies are using Outlook? If internet history tells us anything, Skype is on its way to becoming the AOL of voice and video calling apps. \\
\small\textbf{[Answer Entity]}
\small Microsoft\\
\midrule
\small\textbf{[LLAMA-2-Chat-7B]} \\
\small   Here are the spans from the article that correspond to the query "Companies founded by Bill Gates":

\small * Microsoft\\
\small * Skype \\
\midrule
\small\textbf{[LLAMA-2-Chat-13B]} \\
\small  Here are the spans that correspond to the query "Companies founded by Bill Gates":

* Microsoft, which is owned by Bill Gates. \\
\bottomrule
\end{tabularx}}
\caption{Example of hallucination output of LLAMA-2.}
\label{table:hallucination}
\end{table*}